\documentclass{article}

\usepackage[preprint]{neurips_data_2024}

\usepackage[utf8]{inputenc} 
\usepackage[T1]{fontenc}    
\usepackage[colorlinks,citecolor=lightgray]{hyperref}       
\usepackage{url}            
\usepackage{booktabs}       
\usepackage{amsfonts}       
\usepackage{nicefrac}       
\usepackage{microtype}      
\usepackage{xcolor}         
\usepackage{graphicx}

\usepackage{xspace}
\makeatletter\renewcommand\paragraph{\@startsection{paragraph}{4}{\z@}
	{.25em \@plus1ex \@minus.2ex}{-.5em}{\normalfont\normalsize\bfseries}}\makeatother
\usepackage{times}
\usepackage{epsfig}
\usepackage{graphicx}
\usepackage{amsmath}
\usepackage{amssymb}
\usepackage[linesnumbered,ruled,vlined]{algorithm2e}
\usepackage{caption}
\usepackage{subcaption}
\usepackage{multirow}
\usepackage{eso-pic}
\usepackage[misc]{ifsym}
\usepackage{enumitem}
\usepackage{xspace}
\usepackage{calc}
\usepackage{geometry}
\usepackage[capitalise]{cleveref}
\usepackage{bibunits}
\usepackage{wrapfig}
\usepackage{ulem}
\usepackage{xcolor,colortbl}
\usepackage{mathrsfs}
\usepackage{amsfonts}
\usepackage{bbm}
\usepackage{soul}

\usepackage{fontawesome5}
\pdfmapline{+markerfelt < Markerfelt.ttc <T1-WGL4.enc}

\usepackage{pifont}
\newcommand{\cmark}{\ding{51}}%
\newcommand{\xmark}{\ding{55}}%

\Crefname{equation}{Eqn.}{Eqns.}
\Crefname{section}{Sec.}{Secs.}

\makeatletter
\newcommand{\thickhline}{%
    \noalign {\ifnum 0=`}\fi \hrule height 1pt
    \futurelet \reserved@a \@xhline
}
\makeatother

\makeatletter
\renewcommand{\paragraph}{%
  \@startsection{paragraph}{4}%
  {\z@}{0ex \@plus 0ex \@minus 0ex}{-1em}%
  {\hskip\parindent\normalfont\normalsize\bfseries}%
}
\makeatother

\makeatletter
\renewcommand*{\@fnsymbol}[1]{\ensuremath{\ifcase#1\or *\or \textrm{\Letter}\or \ddagger\or
   \mathsection\or \mathparagraph\or \|\or **\or \dagger\dagger
   \or \ddagger\ddagger \else\@ctrerr\fi}}
\makeatother

\makeatletter
\DeclareRobustCommand\onedot{\futurelet\@let@token\@onedot}
\def\@onedot{\ifx\@let@token.\else.\null\fi\xspace}

 \def\vs{\textit{vs}\onedot}
\def\wrt{w.r.t\onedot} 

\makeatother

\newcommand{\benchtitle}{$\mathcal{V}ideo\mathcal{H}allucer$\xspace}
\newcommand{\bench}{VideoHallucer\xspace}

\newcommand{\highlight}[1]{\cellcolor{blue!25}{#1}}
\newcommand{\highlightse}[1]{\cellcolor{blue!15}{#1}}

\title{\benchtitle: Evaluating Intrinsic and Extrinsic Hallucinations in Large Video-Language Models}

\author{%
  Yuxuan Wang$^{1,2}$ \\ \texttt{wangyuxuan1@bigai.ai} \And Yueqian Wang$^{2,3}$ \\\texttt{wangyueqian@.pku.edu} \And Dongyan Zhao$^{2,3}$ \\\texttt{zhaodongyan@pku.edu} \\
  \AND Cihang Xie$^4$ \\\texttt{cixie@ucsc.edu} \And Zilong Zheng$^{1,2\,\textrm{\Letter}}$\\\texttt{zlzheng@bigai.ai} \\ 
  \AND \normalfont
$^1$ Beijing Institute for General Artificial Intelligence, Beijing China \\
$^2$ State Key Laboratory of General Artificial Intelligence, Beijing, China \\
$^3$ Wangxuan Institute of Computer Technology, Peking University, Beijing, China \\
$^4$ Computer Science and Engineering, University of California, Santa Cruz \\
\AND \url{https://VideoHallucer.github.io}
}

\begin{document}

\maketitle

\begin{abstract}
\label{sec:abstract}

Recent advancements in Multimodal Large Language Models (MLLMs) have extended their capabilities to video understanding. Yet, these models are often plagued by ``hallucinations'', where irrelevant or nonsensical content is generated, deviating from the actual video context. This work introduces \bench, the first comprehensive benchmark for hallucination detection in large video-language models (LVLMs). \bench categorizes hallucinations into two main types: intrinsic and extrinsic, offering further subcategories for detailed analysis, including object-relation, temporal, semantic detail, extrinsic factual, and extrinsic non-factual hallucinations. We adopt an adversarial binary VideoQA method for comprehensive evaluation, where pairs of basic and hallucinated questions are crafted strategically.
By evaluating eleven LVLMs on \bench, we reveal that i) the majority of current models exhibit significant issues with hallucinations; ii) while scaling datasets and parameters improves models' ability to detect basic visual cues and counterfactuals, it provides limited benefit for detecting extrinsic factual hallucinations;
iii)  existing models are more adept at detecting facts than identifying hallucinations. As a byproduct, these analyses further instruct the development of our self-PEP framework, achieving an average of 5.38\% improvement in hallucination resistance across all model architectures. 

\end{abstract}

\begin{figure}[h!]
    \centering
    \vspace{-2em}
    \includegraphics[width=\textwidth]{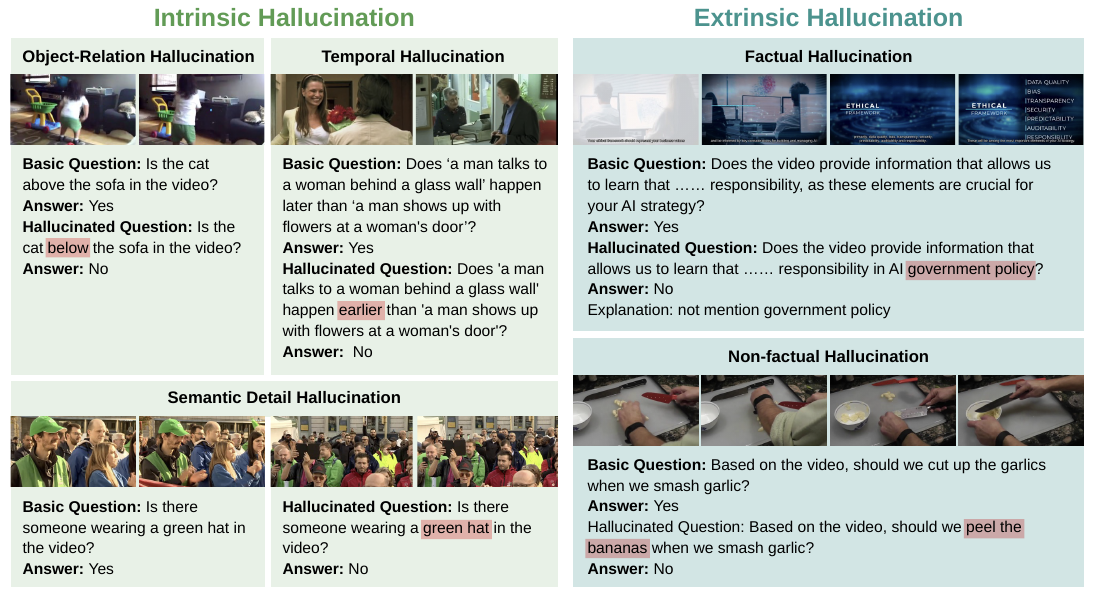}
    \caption{\textbf{Example tasks in \bench.} Each datapoint consists of a basic question, used to test the basic ability of LVLM, and a hallucinated question, containing \colorbox{pink}{hallucinated content} to evaluate the models' ability to detect hallucination. Intrinsic Hallucinations (left) occur when the responses are inconsistent with the original videos. 
    Extrinsic hallucinations (right) happen when the responses cannot be confirmed by the source video. For extrinsic factual hallucination questions,
    we especially add explanation annotation for clarity and further exploration. 
    }
    \label{fig:teaser}
\end{figure}

\section{Introduction}
\label{sec:introduction}


Multimodal Large Language Models (MLLMs) have demonstrated impressive capabilities in both visual understanding and language generation \citep{flamingo, blip2, llava, openai2024gpt4}. However, despite their strong performance on standard benchmarks~\citep{vqa,coco,msvdqa,Xu2016MSRVTTAL,anet}, these models frequently produce incorrect or unsubstantiated responses \wrt visual inputs~\citep{POPE, mmvp, aloha}. This issue, often referred to as ``hallucination'' \citep{chair}, means that \textbf{MLLMs can generate irrelevant or nonsensical content that deviates from the original visual context}. Given these challenges, a natural question arises: How can we examine the vulnerability to hallucinations in MLLMs? Addressing this need not only reveals the extent of hallucination in these models, but also helps identify underlying causes and develop methods to further enhance models.

To address this need, researchers have begun to benchmark hallucinations related to objects~\citep{chair,POPE}, relationships~\citep{correlationqa}, and attributes~\citep{mmhal-bench, amber, mitigating, fohe}, as well as factual information~\citep{bingo, faithscore, easydetect}. However, most existing studies focus on hallucinations involving basic static visual attributes from images in large image-language models. They often overlook potential hallucination issues arising from dynamic content, such as actions, events, and stories, in large video-language models (LVLMs). Furthermore, video-language tasks, such as video summarization~\citep{eduvsum, tvsum}, due to their higher complexity, are in lack of careful evaluation in existing datasets. As a result, we are still unsure of the extent, severity, characteristics, and causes of hallucination issues within LVLMs. Additionally, existing benchmarks focus on certain attributes of hallucinations, lacking a comprehensive and robust evaluation.

To tackle these issues, we introduce \bench, the first comprehensive benchmark designed to assess hallucination in LVLMs. 
Within \bench, we establish a clear taxonomy of hallucinations, distinguishing between two primary categories: intrinsic and extrinsic. Specifically, Intrinsic hallucinations involve generated content that directly contradicts information present in the source video, and can be categorised into three subtypes: object-relation, temporal, and semantic detail hallucinations. While extrinsic hallucinations involve content that cannot be verified from the source, and can be classified as either extrinsic factual, aligning with general knowledge but not present in the source video, or extrinsic non-factual, which includes all the others. 
To avoid confounding factors from LLMs~\citep{POPE,DBLP:journals/corr/abs-2305-13534},
our benchmark focuses on identifying hallucinations in video-language grounding using a binary VQA-based method~\citep{POPE, bingo, mitigating},  Specifically, we introduce an adversarial evaluation\citep{mmvp} with paired questions—one basic and one intentionally hallucinated—to rigorously test models. 
Moreover, we balance 'yes' and 'no' responses to reduce language biases and provide clear explanations to minimize misinterpretation. Comparisons with existing multimodal hallucination benchmark are discussed in \cref{tab:comparison} and \cref{sec:related}.

By comprehensively evaluating twelve LVLMs on \bench, our analysis leads to three significant insights: 
\textbf{First,} our analysis revealed a widespread issue of hallucinations across LVLMs, and more critically, the performance gaps between humans and models in all \bench settings are significant (\cref{sec:result}).
\textbf{Second,} we confirm the benefits brought by scaling on hallucination --- increasing the size of the training dataset or/and the model's parameters improves hallucination detection related to basic visual cues and counterfactuals. 
However, we found that this approach has a limited impact on the models' ability to detect extrinsic factual hallucinations (\cref{sec:result}). 
\textbf{Third,} we observed that current models are more proficient at recognizing facts than detecting hallucinations, where the latter requires the models to discern facts within the context of the source material (\cref{sec:fact-det}). We also found that these shortcomings in recognizing extrinsic factual hallucinations could be partially mitigated by implementing high-quality explanatory mechanisms (\cref{sec:explain}).

Building on the above observations, in \cref{sec:fact-halluc}, we devise  \textbf{Self-PEP}, a plug-and-play framework that bolsters the self-improvement capabilities of models through the integration of explanatory processes. 
By applying Self-PEP, most models demonstrated enhanced performance on the \bench benchmark, with an average improvement of \textbf{5.38\%}. We believe this framework can streamline future research and development in detecting and mitigating hallucinations in video.

\begin{table}[t!]
    \small
    \caption{\textbf{Comparison with existing Vision Hallucination Benchmark} \bench is the first comprehensive benchmark that focuses on hallucination issues in video-language understanding across five different settings. It employs both QA and adversarial formats for robust evaluation.  }
    \label{tab:comparison}
    \centering
    \resizebox{\linewidth}{!}{
    \begin{tabular}{lcc *{6}{p{1cm}} cc}
        \toprule

        \multirow{2}{*}{\textbf{Benchmark}} & \multirow{2}{*}{\textbf{Modality}} & \multirow{2}{*}{\textbf{\# of Ques/Img}} &\multicolumn{6}{c}{\textbf{Hallucination Type}} & \multirow{2}{*}{\textbf{Evaluation}} & \multirow{2}{*}{\textbf{Adversarial}}  \\ 
        \cmidrule{4-7} \cmidrule{8-9} & & & \textbf{Obj.} & \textbf{Rel.} & \textbf{Seman.} & \textbf{Temp.} & \textbf{Fact.} & \textbf{Nonfact.} \\\midrule
        POPE~\citep{POPE} & Image & 3000/500 & \cmark & \xmark & - & - & - & - & BinaryQA & \xmark \\
        HallusionBench~\citep{hallusionbench} & Image, Video (4 frames) & 1129/346 & \cmark & \cmark & \cmark & - & - & - & LLM & \xmark \\
        MMHal-Bench~\citep{mmhal-bench}  & Image& 96/96 & \cmark & \cmark & \cmark & - & - & - & LLM & \xmark \\
        Bingo~\citep{bingo} & Image & 370/370 &  \cmark & - & \cmark & - & - & \cmark & LLM & \cmark \\
        AMBER~\citep{amber} & Image & 15,220/1,004 & \cmark & \cmark & \cmark & - & - & - & BinaryQA/CHAIR & \xmark \\
        EasyDetect~\citep{easydetect} & Image & 420/420 & \cmark & \cmark & \cmark & - & - & \cmark & LLM & \xmark \\
        VHTest~\citep{vhtest} & Image & 1,200/1,200 & \cmark & \cmark & \cmark & - & - & - & LLM & \xmark \\
        PhD~\citep{phd} & Image & 53,976/7,020 &  \cmark & \cmark & \cmark & - & - & \cmark & BinaryQA & \xmark \\ 
        VALOR~\citep{valoreval} & Image & 211/211 & \cmark & \cmark & \cmark & - & - & - & LLM & \xmark \\ \midrule

        \bench (Ours) & Video & 1,800/948 &  \cmark & \cmark & \cmark & \cmark & \cmark & \cmark & BinaryQA & \cmark \\
         \bottomrule
    \end{tabular}}    
    \vspace{-.1in}
\end{table}

\section{Related Work}
\label{sec:related}

\paragraph{Hallucination in NLG}
Generative models in NLP, particularly Large Language Models (LLMs), have shown remarkable proficiency across various language generation tasks. Despite their capabilities, a significant issue persists: the text they produce can sometimes be irrelevant or nonsensical. This issue is known in the field of NLP as ``hallucination.'' Hallucination refers to instances where the content generated by these models does not make sense or deviates from the intended meaning of the source material it is based on~\citep{DBLP:conf/emnlp/Filippova20, DBLP:conf/acl/MaynezNBM20,DBLP:conf/emnlp/ParikhWGFDYD20,DBLP:conf/acl/ZhouNGDGZG21,DBLP:journals/csur/JiLFYSXIBMF23}. The concept of hallucination in NLG tasks can vary slightly, but it is generally categorized into two types based on the relationship between the generated content and the source material. These types are known as Intrinsic Hallucination and Extrinsic Hallucination~\citep{DBLP:conf/emnlp/DziriMZB21, DBLP:journals/corr/abs-2104-14839, DBLP:conf/acl/MaynezNBM20,DBLP:journals/csur/JiLFYSXIBMF23}. Intrinsic hallucination refers to instances where the content generated by a model directly contradicts the information provided in the source material. On the other hand, extrinsic hallucination occurs when the generated content cannot be confirmed or refuted by the source material. In such cases, the content may either align with or contradict extrinsic knowledge. Building on this, \citet{DBLP:conf/acl/CaoDC22} further categorize extrinsic hallucinations into two types: factual and non-factual. Factual hallucinations produce content that can be verified against real-world knowledge, while non-factual hallucinations generate content that is at odds with what is known about the world. Within specialized research domains, there is a divergence of opinion regarding the value of factual hallucinations. Some studies~\citep{DBLP:conf/acl/MaynezNBM20, DBLP:conf/inlg/ThomsonR20}, suggest that factual hallucinations can be beneficial. They argue that the additional knowledge introduced by these hallucinations can enhance the informational quality of the generated content.

\paragraph{Vision Hallucination Benchmark}
Advancements in image-language modeling, particularly in generative tasks such as image captioning and image-based question answering, have led researchers to investigate the phenomenon of hallucination in this field.  The concept of object hallucination in image captioning was first introduced by \citep{chair}. They highlighted the issue of captions erroneously including objects that are absent from the images. To address this, they developed the CHAIR metric, an automatic evaluation tool designed to measure the accuracy of object references in captions by calculating the precision of the hallucinated objects.
Following this development, numerous studies have begun to explore hallucination within image-language models. However, the CHAIR metric itself has come under scrutiny. \citet{POPE} identified certain limitations in CHAIR, noting that the metric's results could be skewed by the way instructions are designed and that the reliance on human-crafted parsing rules could severely restrict its applicability. To overcome these challenges, POPE introduced a binary VQA benchmark specifically tailored for the detection of object hallucination, aiming to provide a more robust and reliable means of evaluation in this area. 
Recent advancements have been made in the development of evaluation toolkits for object hallucination with the aid of LLMs, as evidenced by research \citep{HaELM, cceval, aloha, valoreval}. Concurrently, there is a growing body of work aimed at expanding the concept of object hallucination to include additional visual features. \citep{mmhal-bench, amber, mitigating, fohe, correlationqa} have begun to explore the relationships, attributions, and other visual cues, including counting, OCR and etc. Moreover, \citep{faithscore, hallusionbench, bingo, easydetect, phd}  are pioneering the detection of factual hallucinations within image-language understanding models, marking a significant step forward in the field. Despite the availability of benchmarks for image-language models, there is a notable lack of a clear and comprehensive framework specifically designed to evaluate LVLMs. To address this gap, our work introduces a thorough benchmark dedicated to assessing the performance of LVLMs, with a particular focus on their susceptibility to hallucination. Moreover, \cref{tab:comparison} presents a comparison of \bench with other existing datasets designed for the detection of hallucinations in vision-language models. \bench stands out as the first and most extensive benchmark specifically tailored for LVLM hallucination detection.

\section{The \benchtitle Benchmark}
\label{sec:method}

We will provide a detailed introduction to \bench in the following sections. \cref{sec:construct} will introduce the construction process, including the different types of questions. Then, in \cref{sec:statistic}, we will display the statistical information of \bench. Finally, in \cref{sec:evaluate}, we will demonstrate how we evaluate LVLMs on \bench.

\subsection{Dataset Construction}
\label{sec:construct}

To evaluate hallucination issues in detail, we split our benchmark into intrinsic and extrinsic categories, resulting in five settings. In the following section, we introduce how we construct question-answer pairs for each of these five settings separately. The detailed annotation procedure is discussed in \cref{supp:implement-anno}.

\subsubsection{Intrinsic Hallucination}
\label{sec:intrinsic}

\paragraph{Object-Relation}
\label{sec:obj-rel}

Our work creates the object-relation hallucination setting that concentrates on the objects and their interactions over time, which we organize into three categories: subject, relation, and object. We construct this setting from existing datasets VidOR~\citep{vidor1, vidor2} and VidVRD~\citep{vidvrd}. We follow \citet{POPE} and use templates to generate basic questions about objects and their relations. Annotators then generate semantically distinct yet visually analogous alternatives for these questions to identify hallucinated content. Through this semi-automated annotation process, we produced $400$ question-answer pairs across $183$ videos.

\paragraph{Temporal}
\label{sec:temporal}

To benchmark the temporal hallucination issue in LVLMs, we design a setting to evaluate models' hallucination issues from three dimensions: absolute temporal, relative temporal, and event length.
Utilizing the ActivityNet dataset~\citep{anet}, we create $400$ question-answer pairs spanning $165$ videos. For absolute temporal, we select events located in the first or last of $50$ videos and asking if they occur at the beginning or end. For relative temporal, we choose $75$ pairs of events with clear temporal separation and ask which occurs first. For event length, we compare the durations of $75$ pairs of events, questioning which is longer.

\paragraph{Semantic Detail}
\label{sec:semantic}

Recent research~\citep{mmhal-bench, amber, mitigating, fohe, correlationqa} underscores the importance of hallucinating object attributes in image-language models, such as OCR, object counting, and scene details, which we summarize as "semantic details". To benchmark this in LVLMs, we developed a setting focused on detecting hallucinations related to these details. We've employed a contrastive learning-inspired method using the HawkEye dataset~\citep{hawkeye}, which breaks down long videos into short video segments by semantic similarity. By computing the CLIP score of these video segments, we select segment pairs with a score above $0.85$  as the source video. Annotators then identify semantic differences between these pairs and create basic-hallucinated question-answer pairs from different perspective, resulting in a collection of 400 pairs and corresponding videos.

\subsubsection{Extrinsic Hallucination}
\label{sec:extrinsic}

\paragraph{Factual}
\label{sec:fact}

Our research focuses on detecting extrinsic factual hallucinations—facts consistent with reality but not verifiable from the source material—which can be contextually beneficial or detrimental. In text summarization, hallucinations are unwanted due to the need for source accuracy, while in conversational agents, they may enhance creativity and are more acceptable. We do not assess the overall value of such hallucinations but aim to identify them. At \bench, we curate instructional videos and course lectures, selecting content that stands alone clearly. Specifically, we select videos from YouCook~\citep{youcook}, COIN~\citep{coin}, and EDUVSUM~\citep{eduvsum} 
datasets as our source videos. We prefer shorter videos to minimize complexity. For instructional videos, we analyze tutorial steps and create questions about whether certain steps should be taken to finish a task. To prevent ambiguity, we focus on the final steps to detect hallucinations. We edit videos to exclude these steps and pose similar questions as factual hallucinated questions. For course lectures, annotators summarize the content and create questions to determine if the video contains the summary's content. To create hallucinated questions, annotators modify the summary or add unrelated knowledge, ensuring the summary remains factual but not directly sourced from the video. They also provide explanations for these hallucinated questions to aid further research and improve clarity. As a result, we've created a dataset of $200$ paired basic and hallucinated questions and answers, totaling $400$ pairs from $200$ videos—$150$ instructional videos and $50$ course lectures.

\paragraph{Non-facutal}
\label{sec:nonfact}
Hallucinations that are not based on factual information can be particularly harmful due to their distorted content. Recently, there has been a growing interest among researchers in addressing this issue. At \bench, we specifically develop a setting to aid in the detection of non-factual hallucinations. To ensure consistency with the factual hallucinations, we select the same set of videos and corresponding basic questions used in the factual hallucination setting. For the creation of hallucinated content, we instructed annotators to manually alter the basic questions or infuse them with counterfactual information. The format of these questions remains identical to that used in the factual context. As a result, we created a comprehensive setting consisting of $400$ question-answer pairs linked to $200$ videos.

\subsection{Dataset Statistics}
\label{sec:statistic}

\begin{table}[h]
    \begin{minipage}{0.45\linewidth}
        \centering
            \captionof{table}{\label{tab:statistics} \bench dataset statistics.}
        \resizebox{\linewidth}{!}{%
            \setlength\tabcolsep{6pt}
        \begin{tabular}{lcccccc}
            \toprule
            Statistic & ORH & TH & SDH & EFH & ENFH & All \\
            \midrule
            \# of Ques. & 400 & 400 & 400 & 400 & 400 & 1800\\
            \# of Vids. & 183 & 165 & 400 & 200 & 200 & 948\\
            Avg Ques. len & 23.7 & 69.2 & 27.3 & 92.3 & 94.7 & 61.4\\
            Avg Vids. len (sec.) & 7.0 & 33.8 & 13.5 & 187.0 & 187.0 & 85.6 \\
            \bottomrule
        \end{tabular}
    }
                    \vfill \vskip.1in
                \includegraphics[width=\linewidth]{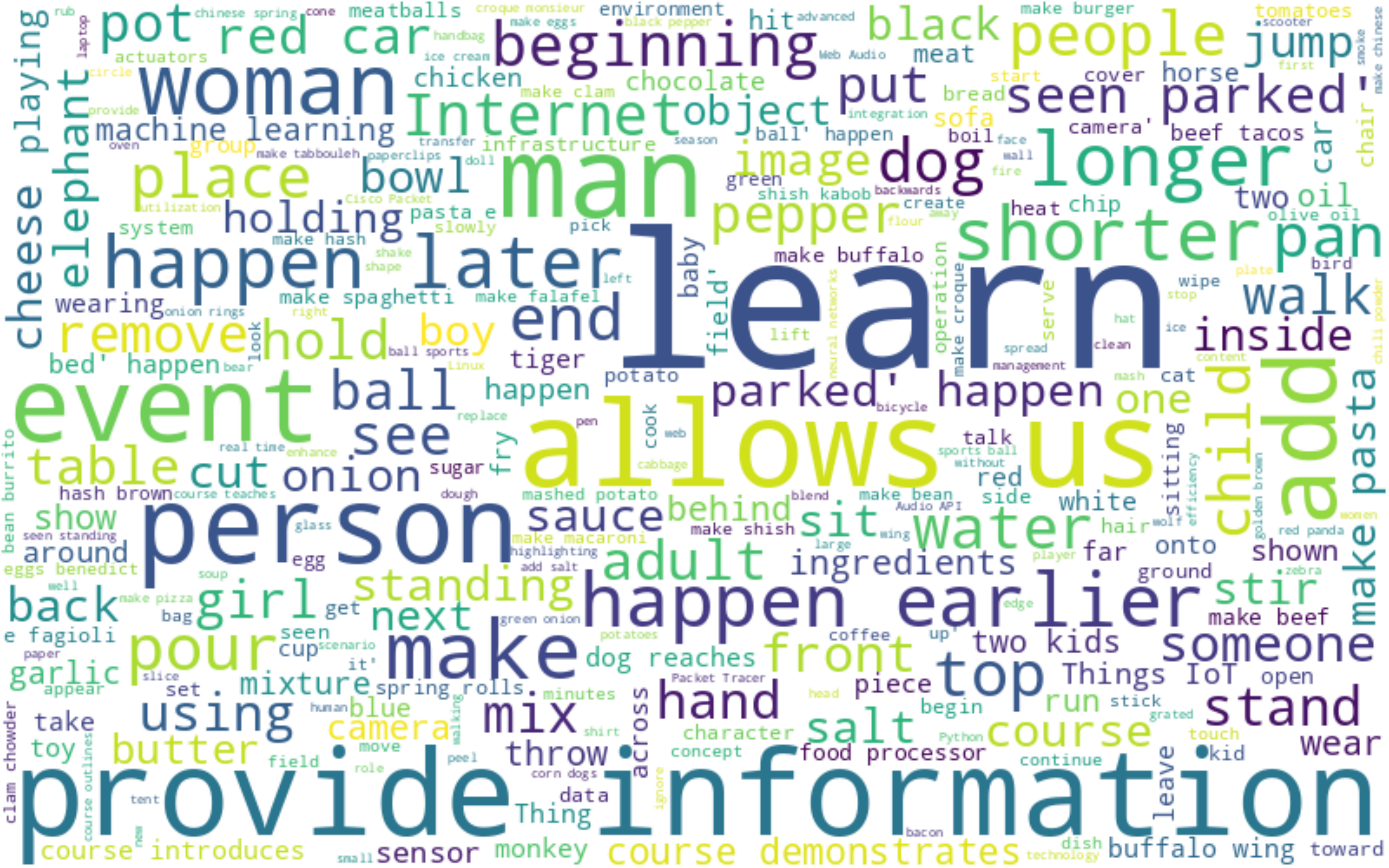}
        \captionof{figure}{\label{fig:word_cloud} Word cloud of questions in \bench.}
    \end{minipage}
    \hfill
    \begin{minipage}{0.5\linewidth}
        \centering
            \includegraphics[width=.9\linewidth]{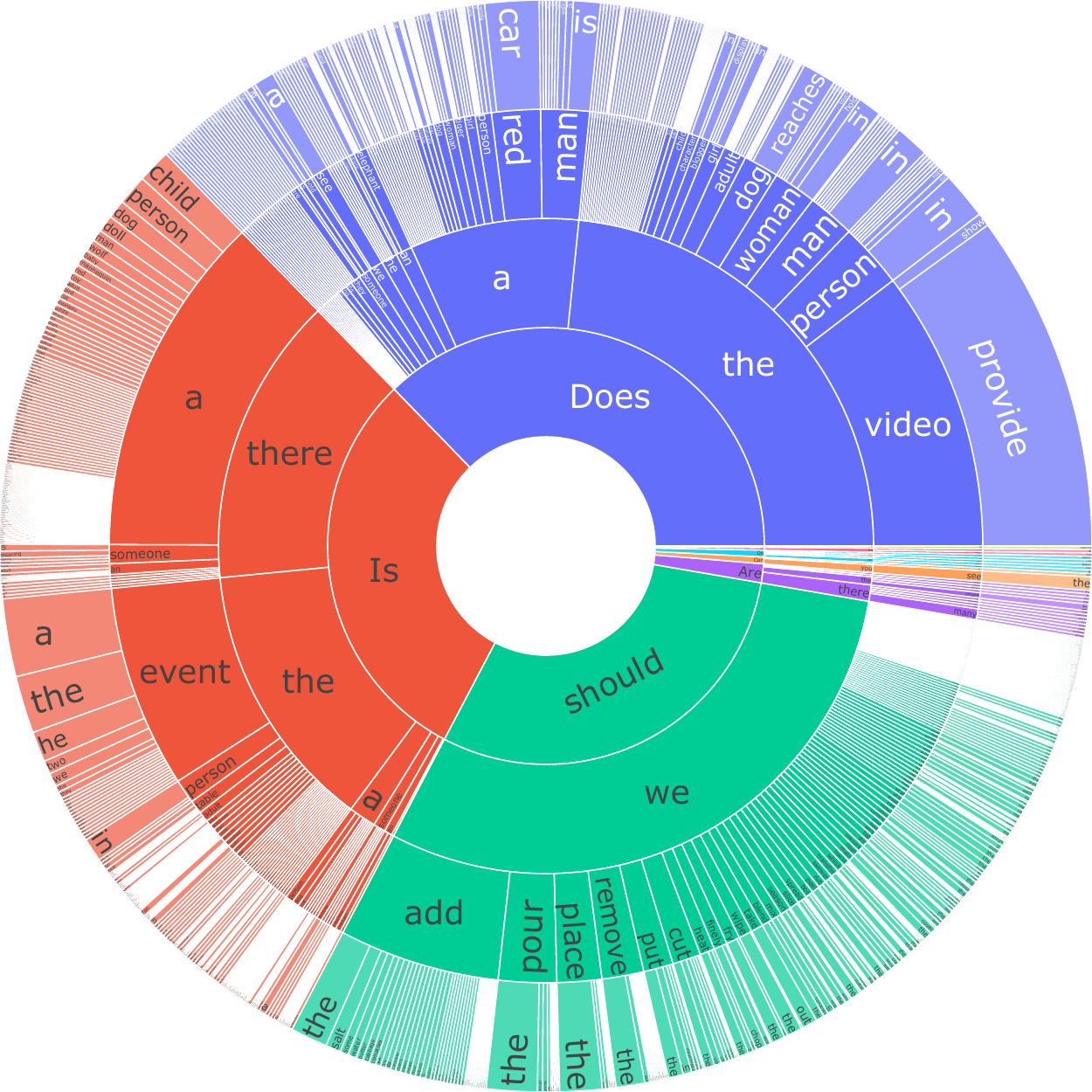}
            \captionof{figure}{\label{fig:question_prefix}Question distribution in \bench}
        \end{minipage}
    \vskip -0.2in
\end{table}

\paragraph{Quantitative Analysis} There are five settings in \bench, each corresponding to a different type of hallucination in video understanding. For each setting, we create $400$ question-answer pairs, including 200 basic questions and $200$ hallucinated questions. To ensure a fair evaluation, the basic questions in both factual and non-factual settings are identical. As a result, we obtain $1,800$ question-answer pairs with an average length of $61.4$ words. Our dataset comprises $948$ videos, ranging from $7$ seconds to $187$ seconds of different settings, with an average length of $85.6$ seconds. These videos encompass most existing common video understanding benchmarks~\citep{msvdqa,Xu2016MSRVTTAL,Jang2017TGIFQATS,Yu2019ActivityNetQAAD} as well as long video benchmarks~\citep{nextqa, egoschema}.

\paragraph{Qualitative Analysis} To present \bench more intuitively, we display the word cloud of our benchmark in ~\cref{fig:word_cloud} and the sunburst chart in ~\cref{fig:question_prefix}. As shown in these figures, the questions within our benchmark are informative and contain three types: ``Does'', ``Is'', and ``Should''. We believe our benchmark can directly and effectively reveal potential hallucination problems within LVLMs.

\subsection{Evaluation}
\label{sec:evaluate}
\paragraph{Hallucination Evaluation}
In this work, we opt for the VQA-based benchmark for the following reasons: (i) Influence of External Factors: Similar to metrics such as BLEU~\citep{bleu} and ROUGE~\citep{rouge}, the values in caption-based benchmarks can be affected by factors such as caption prompt and length~\citep{POPE}.
(ii) Complexity: Methods like CHAIR~\citep{chair} require intricate, human-crafted parsing rules.
(iii) LLM Hallucinations: The potential hallucinations existing in LLMs' generation make it unconvincing to use themselves for self-evaluation~\citep{amber}.
To ensure the credibility of our benchmark, we show a positive correlation between our QA-based evaluation and caption-based methods. More details are discussed in \cref{sec:discussion}.

To mitigate biases such as the distribution of answers and language bias, we develop \bench using an adversarial approach~\citep{mmvp}. Specifically, for each evaluation item, we formulate two types of questions: a \textbf{basic} question and a \textbf{hallucinated} question. The basic question assesses the core capabilities of LVLMs, while the hallucinated question includes deliberately hallucinated content. We then calculate the \textbf{overall} accuracy by considering both the basic and hallucinated questions as a paired set, marking it as a hit only if both questions are answered correctly. We posit that enhancing a model's ability to recognize and counter hallucinations should not compromise its performance on fundamental tasks. This dual-question structure is designed to ensure that improvements in counter-hallucination do not detract from the model's original competencies. 

\paragraph{Bias Evaluation}
In addition to the accuracy, we calculate the \textit{Yes Percentage Difference} (Pct. Diff) and \textit{False Positive Ratio} (FP Ratio) \citep{hallusionbench} to reveal the bias of these LVLMs. Specifically, the \textit{Yes Percentage Difference} is calculated as 
\begin{equation}
    d_y = \frac{|\{M(v, q)=``yes"\}_{(v,q)\in V}|-|\{GT(v, q)=``yes"\}_{(v,q)\in V}|}{|V|}
\end{equation}
where $V$ is the set of video question pairs, $M(v, q)$ is the prediction from models, $GT(v, q)$ is the ground truth. A smaller $d_y$ indicates the number of ``yes'' responses from models is closer to the ground truth, revealing less language bias. the \textit{False Positive Ratio} is calculated as 
\begin{equation}. 
    r_{fp} = \frac{|\{M(v, q)=``yes"\}_{(v, q)\in W}|}{|W|}
\end{equation}
where $W$ is the set of wrongly answered video question pairs. $r_{fp}$ demonstrates the percentage of ``yes'' in all wrongly predicted answers. A value closer to $50\%$ indicates less bias from the models.

\begin{table}[h]
    
    \caption{\textbf{Performance comparison of existing LVLMs on \bench} with additional Yes/No bias analysis. To evaluate the accuracy, we present the performance of all these models on basic questions, hallucinated questions, and the overall score. We \textbf{highlight} the Top 2 models of open-source LVLMs. }
    \label{tab:overall}
    \centering
    \resizebox{\linewidth}{!}{
    \begin{tabular}{l|c|ccccc}
        \toprule
        \multicolumn{2}{c}{}  & \multicolumn{2}{c}{\textbf{Yes/No Bias}} & \multicolumn{3}{c}{\textbf{Accuracy on \bench}} \\ 
        \cmidrule{3-4} \cmidrule{5-7} 
        \textbf{Models} & \textbf{Language Model} &  Pct. Diff ($\sim 0$) & FP Ratio ($\sim0.5$)  & Basic $\uparrow$ & Hallucinated $\uparrow$   &  Overall $\uparrow$  \\
        \midrule
        \multicolumn{7}{c}{\textit{Open-source LVLMs}} \\ \midrule
        VideoChatGPT~\citep{videochatgpt} & LLaMA-7B & 0.40 & 0.89 &92.8 & 10.4 & 6.4 \\
         Valley2~\citep{vally} & LLaMA2-7B & -0.07 & 0.29 & 44.4 &11.5 & 2.8  \\ 
         Video-LLaMA2~\citep{videollama} & LLaMA2-7B & 0.36 & 0.84 & 90.9 & 12.7 & 10 \\
         VideoChat2~\citep{videochat} & Vicuna-7B-v0 & -0.24 & 0.15 &29.7 & 25.8 & 7.8 \\
         Video-LLaVA~\citep{videollava} & Vicuna-7B-v1.5 & 0.36 & 0.91 &  \highlight{95.1} & 20.3 & 17.8 \\
         LLaMA-VID~\citep{llamavid} & Vicuna-7B-v1.5 & 0.29 & 0.83 &  89.9 & 26.6 & 21 \\
         VideoLaVIT~\citep{videolavit} & LLaMA2-7B & 0.36 & 0.91 & \highlightse{94.9} & 21.3 & 18.9 \\
         MiniGPT4-Video~\citep{minigpt4video} & Mistral-7B & 0.18 & 0.62 & 79.4 & 28.6 & 22.3 \\ 
         PLLaVA~\citep{xu2024pllava} & Vicuna-7B-1.5 & \highlightse{0.06} & \highlight{0.53} & 75.1 & \highlightse{55.5} & 38.1 \\ 
         LLaVA-NeXT-Video-DPO~\citep{zhang2024llavanextvideo} & Vicuna-7B-1.5 & \highlight{-0.04} & 0.40 & 62.5 & \highlight{60.9} & 32.0  \\ \midrule
         Video-LLaMA2-13B~\citep{videollama} & LLaMA2-13B & 0.36 & 0.79 & 88.3 & 3.8 & 3.3 \\
         LLaMA-VID-13B~\citep{llamavid} & Vicuna-13B-v1.5 & 0.21 & 0.72 &  85.2 & 36.9 & 29.2 \\ 
         PLLaVA-13B~\citep{xu2024pllava} & Vicuna-13B-1.5 & 0.17 & 0.72 & 87.5 & 48.6 & \highlightse{41.2} \\
         PLLaVA-34B~\citep{xu2024pllava} & Yi-34B & 0.18 & 0.78 & 90.8 & 50.8 & \highlight{45} \\ 
         LLaVA-NeXT-Video-DPO-34B~\citep{zhang2024llavanextvideo} & Yi-34B & 0.07 & \highlightse{0.55} & 73.6 & 51.6 & 32.3  \\\midrule
         \multicolumn{7}{c}{\textit{Closed-source LVLMs}} \\ \midrule
         Gemini-1.5-Pro~\citep{gemini} & - & 0.15 & 0.62 &  83.6 & 42.3 & 37.8 \\
         GPT-4o~\citep{gpt4o} & - & -0.02 & 0.43 &  75.1 & 74.2 & 53.3 \\ \bottomrule
         Human & - & 0.02 & 0.42 & 90 & 88.8 & 85\\
         \bottomrule
    \end{tabular}}
\end{table}

\section{Experiment}
\label{sec:experiment}

In this section, we evaluate the most popular LVLMs on our \bench. We first present the setups of these models (\cref{sec:setup}), followed by the main results and performance analysis (\cref{sec:result}). Additionally, we compare the effectiveness of current Image-Language Models on the object-relation and semantic details settings of \bench (\cref{sec:img}). Finally, we reveal the human performance on our benchmark (\cref{sec:human}).



\subsection{Setups}
\label{sec:setup}
We assess twelve LVLMs, comprising ten open-source models (7B unspecified), including VideoChatGPT, Valley2, Video-LLaMA2, VideoChat2, Video-LLaVA, LLaMA-VID, VideoLaVIT, MiniGPT4-Video, PLLaVA, and LLaVA-NeXT-Video-DPO, and two closed-source models,  Gemini-1.5-Pro and GPT-4o. To make a fair comparison, we set all these baselines following their original setting including the number of frames and generation hyper-parameters.

\subsection{Main Benchmark Results}
\label{sec:result}

The overall results are delineated in \cref{tab:overall}. We find that although all models demonstrate strong capabilities in answering basic questions, they experience a significant decline in accuracy when confronted with hallucinated questions. The overall accuracy significantly drops compared to the accuracy on basic and hallucinated questions due to a mistake in one of the question pairs. This pattern implies a widespread susceptibility to hallucination issues among the current models. Regarding the ``Yes/No Bias'', models with an obvious bias are more likely to have hallucination problems. Specifically, we find that most models tend to generate ``Yes'' answers, except for VideoChat2, which is more likely to generate ``No'' answers. Since PLLaVA and VideoChat2 share the same video tuning data, the difference stems from the image data. Therefore, we believe the bias and hallucination issues in VideoChat2 originate from the training image data. Additionally, LLaVA-NeXT-Video-DPO shares similar tuning data with Video-LLaVA, and as a result, the DPO strategy significantly reduces bias and hallucination. Generally, we do not find a huge gap between open-source models and closed-source models, but all models lack significant human-like hallucination detection capabilities.

\begin{figure}[!h]
  \centering
  \small
  \includegraphics[width=\textwidth]{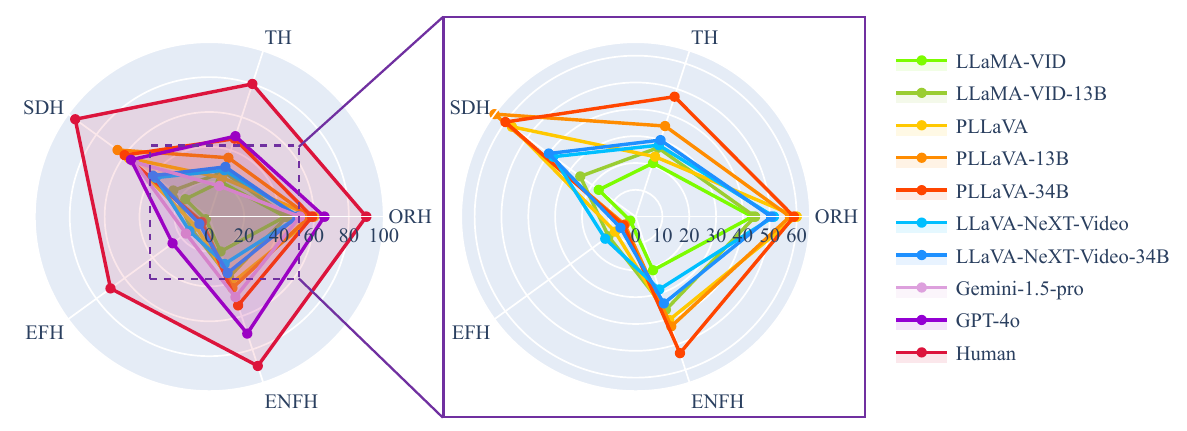}
  \caption{\textbf{Comparative analysis of models using \bench across various settings.} The left displays the complete set of results, while the right provides a magnified view to facilitate a closer examination of the performance details among the open-source models of different scales.}
  \label{fig:detail_radar}
\end{figure}

Here, we will dig deeper into different types of hallucinations in LVLMs. We illustrated the more detailed analysis in Figure~\ref{fig:detail_radar}. We have the following findings. 

First, when comparing the various dimensions of the radar chart, we observe that \ul{most models exhibit fewer hallucinations in the object-relation setting (ORH) than in other areas}. Specifically, the performance of most models in this setting is centered around $50\%$.  This observation points to a deficiency in existing modeling techniques to detect hallucinations beyond elementary visual cues. 

Second, regarding semantic detail hallucination (SDH), when comparing LLaMA-VID, LLaVA-NeXT-Video, and PLLaVA with training data increasing, we find \ul{models with more training data significantly perform better than others}. Moreover, we compare the different scales of these models on non-factual settings (ENFH), and we find larger models markedly surpass smaller models. We ascribe this superiority to the effects of data and model parameter scaling, which appear to bolster the model's prowess in discerning visual details and retaining world knowledge. 

Finally, for extrinsic factual hallucination (EFH), most models demonstrate inability in this setting. To be more specific, \ul{most existing models can't discern hallucination issues that align with world knowledge but contradict the video context}. Given the significant differences in model performance in the two settings of extrinsic hallucinations, we will explore this issue in greater depth in the \cref{sec:fact-halluc} to uncover the reasons behind this phenomenon.

\subsection{Comparison between Image-Language Models}
\label{sec:img}

    

\begin{table}[!htbp]
    \caption{\textbf{Comparison between Video-langauge models and Image-language models}. We present the results for basic questions, hallucinated questions, and the overall score. We highlight the Top 2 model of accuracy on \bench.}
    \label{tab:image-language}
    \centering
    \resizebox{\linewidth}{!}{
    \begin{tabular}{lcccccc}
        \toprule
        \multicolumn{1}{c}{} & \multicolumn{3}{c}{\textbf{Object-Relation}} & \multicolumn{3}{c}{\textbf{Semantic Detail}} \\
        \cmidrule{2-4} \cmidrule{5-7}
        \textbf{Models} & Basic & Hallucinated & Overall & Basic & Hallucinated & Overall \\\midrule
        \multicolumn{7}{c}{\textit{Video-Language Models}}\\ \midrule
         LLaMA-VID~\citep{llamavid} & 78.5 & 59 &  43.5 & \highlight{89} & 24 & 17 \\ 
         MiniGPT4-Video~\citep{minigpt4video} & 80.5 & 34.5 &   27.5 & 78.5 & 27.5 & 23.5 \\
         PLLaVA~\citep{xu2024pllava} & 76 & 76.5 &   60 & 83 & 71.5 & \highlight{57} \\
         LLaVA-NeXT-Video-DPO~\citep{zhang2024llavanextvideo} & 72 & 73 &   51.5 & 63.5 & 69 & 38\\
         Gemini-1.5-Pro~\citep{gemini} & \highlight{84.5} & 56 &  52 & \highlight{89} & 63 & 53.5 \\ 
         GPT-4o~\citep{gpt4o} & \highlightse{81} & \highlight{82.5} & \highlight{66} & 63 & \highlight{87.5} & \highlightse{55.5} \\ 
         \midrule
         \multicolumn{7}{c}{\textit{Image-Language Models}}\\ \midrule
         LLaVA-1.5~\citep{llava} & 79.5 & 79 & \highlightse{61.5} & 75.5 & 63.5 & 43 \\
         GPT4V~\citep{openai2024gpt4} & 65 & \highlightse{81.5} & 55 & 59 & \highlightse{87} & 49.5 \\
         \bottomrule
    \end{tabular}}
\end{table}

We carried out a comparative study of two types of multimodal models: image-language models and video-language models. For this analysis, we selected the top-performing LLaVA-1.5 and GPT4V. To facilitate a fair comparison, we used the middle frame from each video as the input for the image-language models. The findings, presented in Table ~\ref{tab:image-language}, show that open-source image-language models have a superior performance in detecting object-relation hallucination, even though the dataset 
includes dynamic interactions. We believe this discrepancy in performance is due to two main reasons. First, there is a significantly larger amount of training data 
available for images than for videos. This means that image-language models benefit more from the scaling law. Second, videos tend to include more noise than images, making them more likely to encounter the hallucination problem. In light of these insights, we suggest that future development of LVLMs could be enhanced by integrating image datasets~\citep{videollava} or by building upon the foundations of existing image-language models~\citep{zhang2024llavanextvideo, xu2024pllava}.

\subsection{Human Evaluation}
\label{sec:human}

We recruited three individuals proficient in English to evaluate the \bench benchmark. Each evaluator possesses basic computer knowledge, which is essential since part of the extrinsic dataset includes computer science courses. To mitigate potential bias from the evaluators, we randomized the sequence of question-answer pairs to ensure that basic and hallucination pairs did not appear consecutively. We then computed the Pearson correlation coefficient among all evaluators' scores, which resulted in a moderate agreement with a value of $0.557$.

\section{Self-PEP: Self-improvement with Predict-Explain-Predict}
\label{sec:fact-halluc}

In \cref{sec:result}, we find that most models perform worse in the factual hallucination setting compared to others on \bench. In this section, we aim to understand the reasons behind this by comparing the fact detection and hallucination detection abilities of these models (\cref{sec:fact-det}). Based on our findings, we propose a simple yet effective method with explanation to mitigate the models' hallucination issues (\cref{sec:explain}).

\subsection{Fact Detection vs. Hallucination Detection}
\label{sec:fact-det}



One of our key findings suggests that these models struggle to identify external factual hallucinations within video content. As a result, we have redirected our focus in this section to assess the models' capability to discern factual knowledge. To this end, we crafted additional QA pairs derived from questions in the extrinsic 
hallucination setting. These questions aim to determine the factual accuracy of statements in the original extrinsic hallucination questions.

\begin{wrapfigure}{r}{.5\linewidth}
    \centering
    \includegraphics[width=\linewidth]{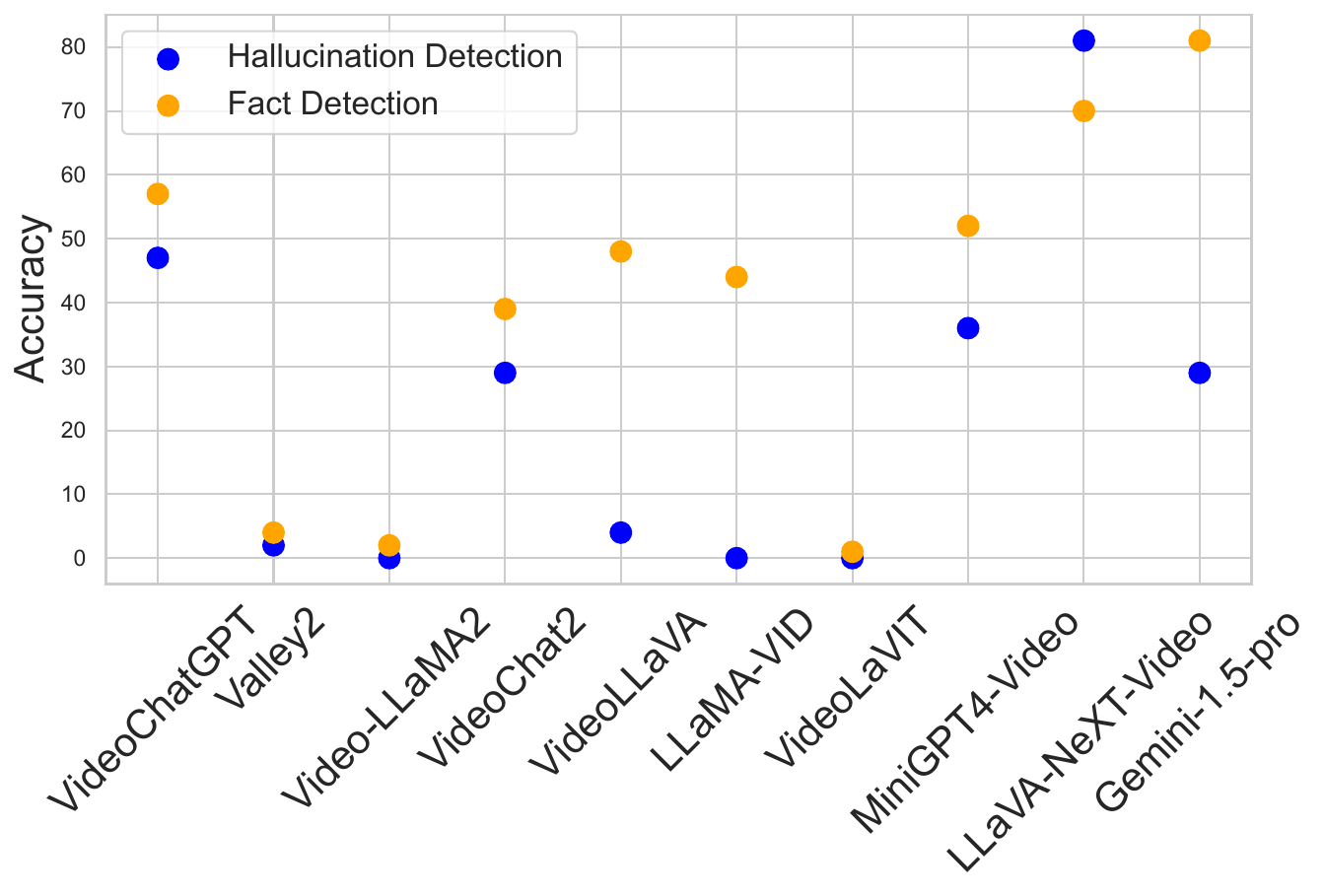}
    \caption{\textbf{Results Comparison of Hallucination Detection and Fact Detection for Extrinsic Hallucination}. Most models are more adept at detecting factuality than detecting hallucination.}
    \label{fig:fact-det}
    \vspace{-.1in}
\end{wrapfigure}
In our experiment, we utilized course videos from this context due to their rich factual content. For instance, we paraphrased extrinsic factual hallucination questions to ``Does the following course summary include any non-factual information? \{summary\}'', and non-factual hallucination questions to ``Does the following course summary encompass all essential factual information? \{summary\}''. By setting the correct answer to these questions as "no", we sought to counteract any language biases present in language models. The experimental results, illustrated in \cref{fig:fact-det}, reveal that most models are more adept at detecting factual knowledge than at detecting hallucinations. This indicates that current methods can effectively recognize counterfactual content. However, their ability to detect hallucinations in video data is markedly limited, highlighting a significant potential for improvement in this area. Among these models, we find the LLaVA-NeXT-Video-DPO performs better on hallucination detection than fact detection, we believe this gain comes from the DPO. Therefore, we believe that human feedback could mitigate hallucination issues in LVLMs.



\subsection{Self-PEP Framework}
\label{sec:explain}


\begin{figure}[ht!]
\begin{minipage}{.49\linewidth}
\centering
    \includegraphics[width=\linewidth]{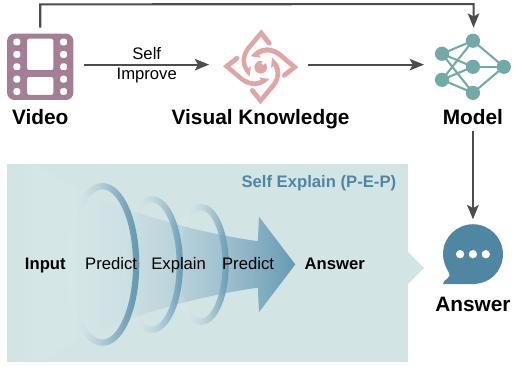}
    \caption{\textbf{The Self-PEP Framework}. 
    The self-improvement strategy (top-row) leverages extracted visual knowledge to introduce related information visibly. The self-explanation strategy (bottom-row) employs a predict-explain-predict scheme to mitigate hallucination issues with self-generated explanations.}
    \label{fig:self-pep}
\end{minipage}
\hfill
\begin{minipage}{.49\linewidth}
    \includegraphics[width=\linewidth]{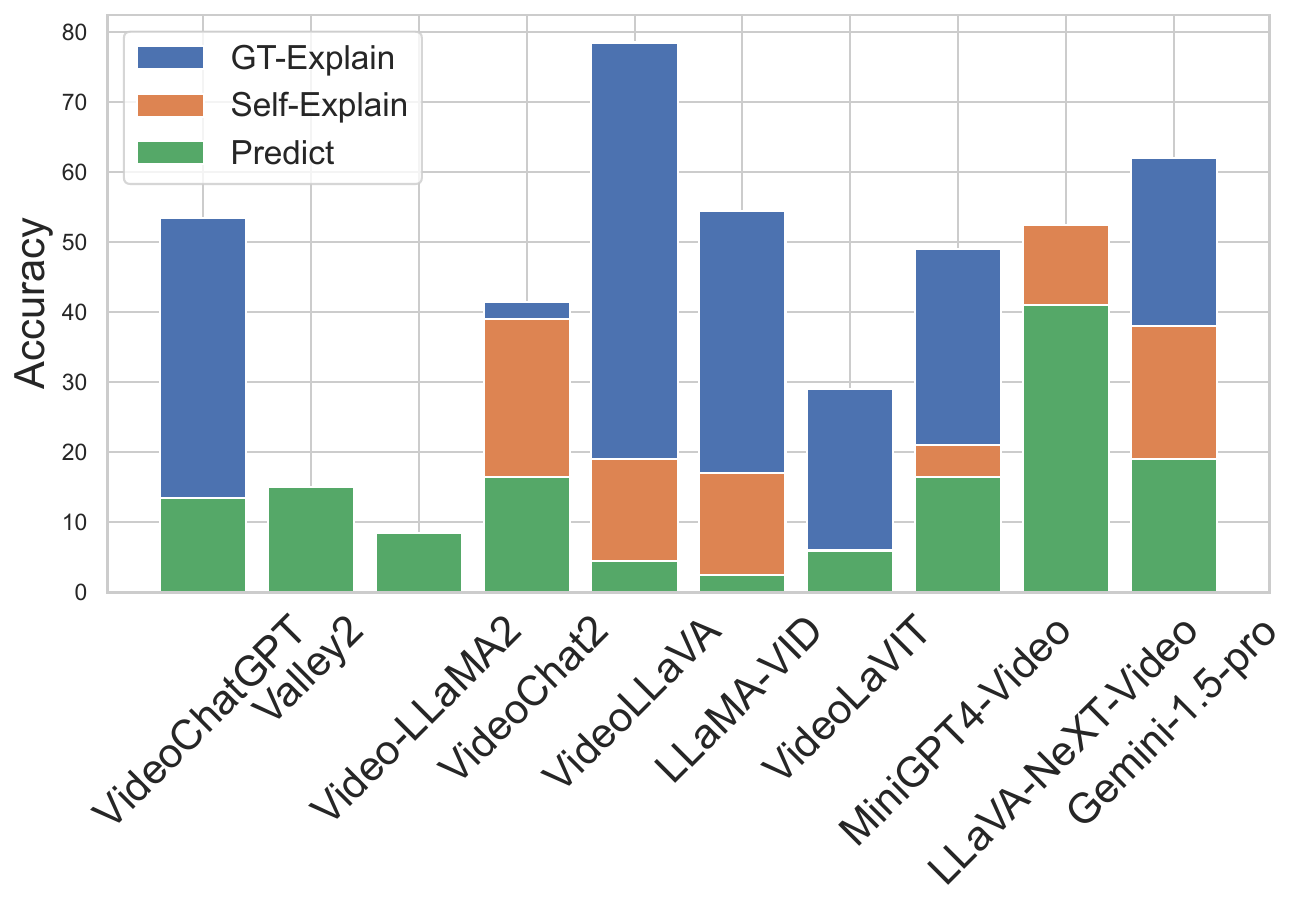}
    \caption{\textbf{Results of Self-explain Strategy on Extrinsic Factual Hallucination}. Most models could benefit from the self-explain strategy, and the ground-truth explanation would improve models' performance on extrinsic hallucination significantly. }
    \label{fig:PEP_res}
\end{minipage}
\end{figure}

Advances in LLMs have seen significant improvements through costly human feedback~\citep{rlhf, mmhal-bench}. To reduce expenses, recent studies~\citep{Madaan2023SelfRefineIR,Shinn2023ReflexionLA,selfee2023,Yan2023LearningTS,Pan2023AutomaticallyCL,Chen2023SeeTC,Zhang2023MultimodalCR,you2023idealgpt,Lightman2023LetsVS} focus on self-improvement of LLMs for better performance and clarity. Our research in \cref{sec:fact-det} shows LLMs are better at identifying facts than hallucination, with the latter demanding a firm grasp of factual context.
Building on these insights, we investigate further in this section. We examine how explanation can affect a model's ability to detect hallucinations. Our methodology involves prompting the model to generate a prediction and then provide an explanation for its output. We then use the explanation to refine the model's predictive accuracy. This experiment is conducted within the context of extrinsic factual hallucination detection, and the results are illustrated in \cref{fig:PEP_res}. Although the self-generated explanations are less impactful compared to those derived from ground truth explanations, our results demonstrate that they do contribute to improving the model's capability in hallucination detection. This improvement is related to the model's proficiency in fact detection.


\begin{table}[!t]
    
    \caption{\textbf{Results of Self-PEP Framework}. We \textbf{highlight} the best performing model after applying our framework. For a clear comparison, we \textcolor{gray}{annotate} all the models after applying our framework. }
    \label{tab:improvement}
    \small
    \centering
    \resizebox{\linewidth}{!}{
    \begin{tabular}{lccccccc}
        \toprule
        \textbf{Models} &  \textbf{Object-Relation} &\textbf{Temporal} & \textbf{Semantic Detail} & \textbf{Factual} & \textbf{Non-factual} & \textbf{Overall Accuracy}  \\\midrule
        VideoChatGPT~\citep{videochatgpt}  & 6 & 0 & 2 & 7 & 17 & 6.4 \\
        \rowcolor{lightgray!50}
        \textsl{\tiny{+Self-PEP}} & 33.5\textsl{\tiny{+27.5}} & 4.5\textsl{\tiny{4.5}} & 22.5\textsl{\tiny{+20.5}} & 14\textsl{\tiny{+7}} & 30\textsl{\tiny{+13.0}} & 20.9\textsl{\tiny{+14.5}}  \\
        Valley2~\citep{vally}  & 4.5 & 3 & 2.5 & 0.5 & 3.5 & 2.8 \\
        \rowcolor{lightgray!50}
        \textsl{\tiny{+Self-PEP}} & 10\textsl{\tiny{+5.5}} & 5.5\textsl{\tiny{+2.5}} & 1.5\textsl{\tiny{-1}} & 1.5\textsl{\tiny{+1}} & 4.5\textsl{\tiny{+1}} & 4.6\textsl{\tiny{+1.8}}  \\
        Video-LLaMA-2~\citep{videollama}  & 18 & 7.5 & 1 & 6.5 & 17 & 10 \\
        \rowcolor{lightgray!50}
        \textsl{\tiny{+Self-PEP}} & 12\textsl{\tiny{-6}} & 5.5\textsl{\tiny{-2}} & 8\textsl{\tiny{+7}} & 15.5\textsl{\tiny{+9}} & 26\textsl{\tiny{+9}} & 13.4\textsl{\tiny{+3.4}}  \\
        VideoChat2~\citep{videochat}  & 10.5 & 7.5 & 9 & 7 & 5 & 7.8 \\
        \rowcolor{lightgray!50}
        \textsl{\tiny{+Self-PEP}} & 34\textsl{\tiny{+23.5}} & 15\textsl{\tiny{+7.5}}& 27.5\textsl{\tiny{+18.5}} & 19.5\textsl{\tiny{+12.5}} & 21.5\textsl{\tiny{+16.5}}& 23.5\textsl{\tiny{+15.7}} \\
        VideoLLaVA~\citep{videollava}  & 34.5 & 13.5 & 12 & 3 & 26 & 17.8 \\
        \rowcolor{lightgray!50}
        \textsl{\tiny{+Self-PEP}} &52\textsl{\tiny{+17.5}} & 5.5\textsl{\tiny{-8}} & 36\textsl{\tiny{+24}} & 11\textsl{\tiny{+8}} & 34\textsl{\tiny{+8}} & 27.7\textsl{\tiny{+10}}  \\
        LLaMA-VID~\citep{llamavid}  & 43.5 & 21 & 17 & 2.5 & 21 & 21 \\
        \rowcolor{lightgray!50}
        \textsl{\tiny{+Self-PEP}} & 44.5\textsl{\tiny{+1}} & 14\textsl{\tiny{-7}} &36.5\textsl{\tiny{+19.5}} & 22\textsl{\tiny{+19.5}} & 33\textsl{\tiny{+12}} & 30\textsl{\tiny{+9}}  \\
        VideoLaVIT~\citep{videolavit}  & 35.5 & 25.5 & 10.5 & 4 & 19 & 18.9 \\
        \rowcolor{lightgray!50}
        \textsl{\tiny{+Self-PEP}} & 24\textsl{\tiny{-11.5}} & 0.5\textsl{\tiny{-25}} & 22\textsl{\tiny{+11.5}} & 6\textsl{\tiny{+1}} & 13\textsl{\tiny{-6}} & 13\textsl{\tiny{-6}}  \\
        MiniGPT4-Video~\citep{minigpt4video}  & 27.5 & 18 & 23.5 & 12 & 30.5 & 22.3 \\
        \rowcolor{lightgray!50}
        \textsl{\tiny{+Self-PEP}} & 47\textsl{\tiny{+19.5}} & 21.5\textsl{\tiny{+3.5}} & 34.5\textsl{\tiny{+11}} & 14\textsl{\tiny{+2}} & 35.5\textsl{\tiny{+5}} & 30.5\textsl{\tiny{+8.2}} \\ 
        PLLaVA~\citep{xu2024pllava}  & 60 & 23.5 & 57 & 9.5 & 40.5 & 38.1 \\
        \rowcolor{lightgray!50}
        \textsl{\tiny{+Self-PEP}} & 52\textsl{\tiny{-8}} & 7\textsl{\tiny{-15.5}} & 46.5\textsl{\tiny{-10.5}} & 16.5\textsl{\tiny{+7}} & 36.5\textsl{\tiny{-1}} & 31.7\textsl{\tiny{-5.6}} \\
        LLaVA-NeXT-Video-DPO~\citep{zhang2024llavanextvideo}  & 51.5 & 28 & 38 & 14 & 28.5 & 32 \\
        \rowcolor{lightgray!50}
        \textsl{\tiny{+Self-PEP}} & 51\textsl{\tiny{-0.5}} & 15\textsl{\tiny{-13}} & 37.5\textsl{\tiny{-0.5}} & 11\textsl{\tiny{-3}} & 16\textsl{\tiny{-12.5}} & 26.1\textsl{\tiny{-5.9}} \\ \midrule
        Gemini-1.5-Pro~\citep{gemini}  & 52 & 18.5 & 53.5 & 16.5 & 48.5 & 37.8 \\
        \rowcolor{lightgray!50}
        \textsl{\tiny{+Self-PEP}} & \textbf{56}\textsl{\tiny{+4}} & \textbf{44}\textsl{\tiny{+25.5}} & \textbf{64}\textsl{\tiny{+10.5}} & \textbf{33}\textsl{\tiny{+16.5}} & \textbf{63}\textsl{\tiny{+14.5}} & \textbf{52}\textsl{\tiny{+14.2}}  \\
         \bottomrule
    \end{tabular}
    }
\end{table}

Building on these insights, as depicted in \cref{fig:self-pep}, we devise an innovative framework called \textbf{Self}-Improvement with \textbf{P}redict-\textbf{E}xplain-\textbf{P}redict (\textbf{Self-PEP}). This framework is designed to enhance the model's resilience against the tendency to produce hallucinations. It capitalizes on the model's established strength in fact detection over hallucination identification. The Self-PEP framework operates in two phases: self-improvement and self-explanation. In the self-improvement phase, the model autonomously extracts visual knowledge~\citep{woodpecker}, while the self-explanation phase involves a three-step process: predict, then explain, and finally refine the prediction using the explanation. The implementation details of the Self-PEP framework and qualitative results are provided in \textbf{Appendix \ref{supp:implement-selfpep}}. 

The efficacy of this framework is demonstrated in \cref{tab:improvement}, where we note that Self-PEP significantly boosts the performance of most models on the \bench benchmark, yielding substantial improvements. In addition, we find the improvements on hallucinated questions are much more significant than the basic questions. When comparing all different settings, we find our methods could consistently improve all models' performance on extrinsic factual hallucination. Our method, for instance, could potentially harm PLLaVA in other settings while improving its performance solely on the factual setting. Consequently, we believe our method could reduce hallucination issues within existing LVLMs, particularly factual hallucinations, which are consistently present in LVLMs.


\section{Conclusion and Discussion}\label{sec:discussion}

In this work, we introduce \bench, a novel and comprehensive benchmark for detecting hallucinations in LVLMs. Our adversarial approach, which challenges models with paired questions, ensures a thorough evaluation of a model's ability to discern hallucinations. Through rigorous testing of 12 LVLMs, we have identified the pervasive nature of spurious hallucinations and the limitations of model scaling in addressing certain types of hallucinations. Our innovative Self-PEP framework has demonstrated the potential to significantly enhance model performance against hallucinations.



\paragraph{Hallucination \vs Adversarial Attacks}
Adversarial attacks are deliberately crafted inputs designed to provoke erroneous outputs from a model, potentially leading to various security breaches such as unauthorized data access, fraud, system intrusion, malware deployment, content manipulation, and service disruption. In contrast, our tool, \bench, is specifically tailored to assess perceptual challenges. It evaluates whether a model can produce accurate and reliable responses based on the provided context, without being misled by superficially plausible but incorrect information.

\paragraph{\bench \vs Other Video-Language Benchmarks}
\bench is primarily designed to evaluate issues related to hallucinated content and the faithfulness of generated text from LVLM. This approach is relatively direct and specific. In comparison, other benchmarks may concentrate on basic visual understanding or more abstract cognitive tasks such as logical reasoning or advanced scene comprehension.

\paragraph{VQA-based benchmark \vs Caption-based benchmark}
Existing hallucination benchmarks primarily encompass two types of benchmarks and evaluation pipelines found in existing research: VQA-based and Caption-based. The VQA-based benchmark, exemplified by methods like POPE~\citep{POPE}, employs binary question-answering to detect instances of object hallucination. In contrast, caption-based benchmarks~\citep{chair, HaELM, cceval} assess the precision or recall of hallucinated objects within captions, with evaluations conducted either through rule-based parsing or LLMs. In our study, we opt for the VQA-based benchmark for several reasons. Firstly, akin to established evaluation metrics for content generation such as BLEU~\citep{bleu} and ROUGE~\citep{rouge}, these values can be influenced by extrinsic factors, including the nature of the caption prompt and the caption's length~\citep{POPE}. Secondly, methods like CHAIR~\citep{chair} depend heavily on intricate, human-crafted rules for parsing, which increases complexity. Thirdly, although recent advancements have incorporated LLMs for evaluation, employing a tool that itself is prone to have hallucination issues for assessment purposes is problematic. To establish the credibility of our benchmark, we demonstrate a positive correlation between our QA-based evaluation and the caption-based methods. Further details on this analysis are available in \textbf{Appendix \ref{supp:evaluation-analysis}}.



\clearpage
\bibliographystyle{unsrtnat}
\bibliography{reference}






\clearpage

\clearpage

\appendix

\section{Appendix}

We provide appendices and supplementary materials as follows:

\begin{itemize}[noitemsep, leftmargin=25pt, topsep=2pt]
    \item In Section \ref{supp:evaluation-analysis}, we show the detailed analysis of two ways of evaluation methods for vision hallucination benchmark.
    \item Section \ref{supp:implement-anno} demonstrates the detailed annotation procedures.
    \item Section \ref{supp:implement} outlines the implementation details of our study
    \begin{itemize}
        \item In \ref{supp:implement-setup}, we show baseline configurations
        \item In \ref{supp:implement-eval}, we demonstrate the evaluation prompts
        \item In \ref{supp:implement-selfpep}, we reveal the implementation details of Self-PEP
    \end{itemize}
    \item  Section \ref{supp:exp-analysis}, we show the analysis of the coherence of the self-generated explanation and the groud-truth explanation.
    \item \ref{supp:detail-quant}. Detailed Quantitative Results
    \begin{itemize}
        \item In Section \ref{supp:overall}, we present the comprehensive results for the \bench benchmark.
        \item Section \ref{supp:self-pep} details the performance of Self-PEP on \bench
    \end{itemize}
    \item \ref{supp:example}. Example Questions. We provide examples from \bench that cover various types of hallucinations: object, spatial relation, temporal relation, absolute temporal, relative temporal, and semantic detail (including attribution, event, count, OCR, camera, and scene), as well as extrinsic factual (instruction, course) and non-factual (instruction, course) hallucinations.
    \item We show the limitations and ethic states in ~\cref{supp:limitation}.
\end{itemize}

\subsection{Evaluation Method Analysis}
\label{supp:evaluation-analysis}

\begin{wrapfigure}{r}{.5\linewidth}
    \centering
    \includegraphics[width=\linewidth]{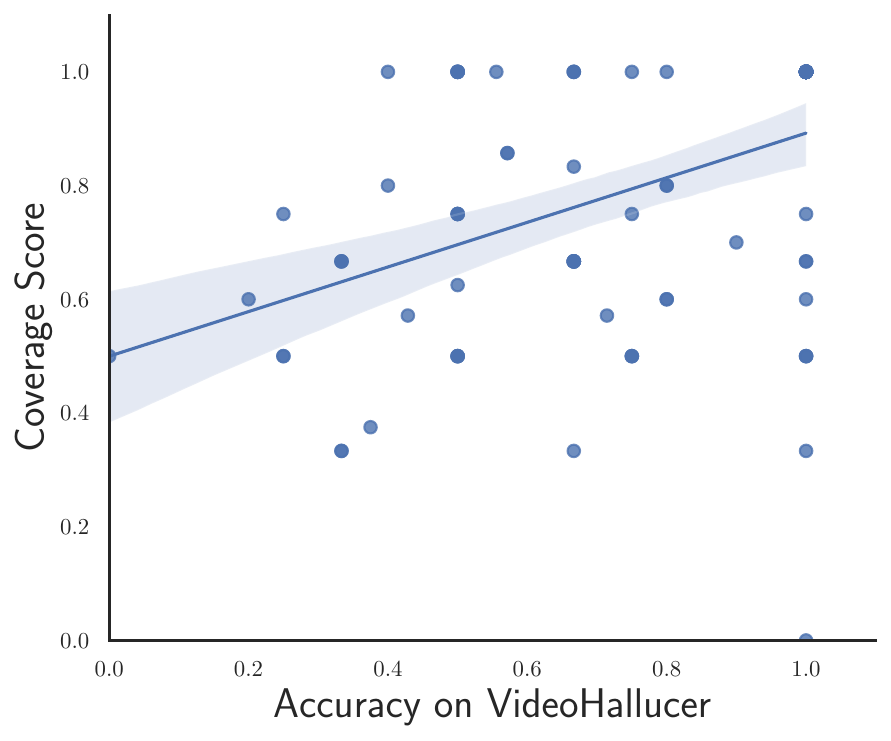}
    \caption{Correlation of Accuracy on \bench and Coverage score}
    \label{fig:correl}
\end{wrapfigure}

In this study, we investigate the relationship between \bench (QA-based mehtod) and the caption-based method. To determine the effectiveness of our QA-based method in comparison to the caption-based approach, we use the Pearson correlation coefficient to measure the coherence. Our focus is on two well-established caption-based evaluation metrics: the CHAIR score~\citep{chair} and the Coverage Score~\citep{cceval}. Specifically, we randomly select a sample of 100 images from the COCO dataset to investigate the correlation between our newly introduced accuracy on \bench and the traditional metrics employed for image caption. We apply the experiments on LLaVA-1.5. To prevnet the issue of hallucination in LLMs, we implement a rule-based method to calculate the Coverage Score similar to CHAIR. Our findings reveal a moderate positive correlation with the Coverage Score ($r = 0.477$) and a weak negative correlation with the CHAIR score ($r = -0.139$). We illustrate the correlation between the overall accuracy on \bench and the Coverage Score in ~\cref{fig:correl}. These results indicate that overall accuracy on \bench correlates positively with the caption-based methods, suggesting that \bench is a reliable and robust alternative to caption-based evaluation methods for assessing hallucinations in generative models.

\subsection{Annotation Details}
\label{supp:implement-anno}
To enhance the quality and reduce the annotation costs of our benchmark, we have developed a semi-automatic pipeline for constructing \bench. The process begins with the use of readily available tools and templates to create or collect initial pairs of basic and hallucinated content. Subsequently, we engage annotators to meticulously review and refine the question-answer pairs, ensuring they are well-grounded in the video content. This review focuses on three key aspects: correctness, ambiguity, and fluency. To guarantee the high standard of \bench, we implement a two-stage human verification process. 
In the following paragraphs, we will give a detailed introduction to different sets of \bench.

\paragraph{Object-Relation Hallucination}
In the domain of vision-language modeling, object hallucination is a persistent challenge, particularly within the context of image captioning tasks. These tasks demand that the model accurately identifies and describes objects in an image. A significant body of research has been dedicated to addressing this issue, encompassing various approaches such as the development of benchmarks, the refinement of evaluation metrics, and the exploration of both analytical and practical solutions. More recently, the focus of research has expanded from object hallucination to include relation hallucination, with a particular emphasis on spatial relationships between objects. Building on this foundation, our work takes a novel step by applying these concepts to video content. We aim to capture not only the objects but also the dynamic interactions between them, considering both spatial and temporal dimensions. We posit that the ability to detect basic visual concepts is a fundamental skill for LVLMs.  

We posit that objects and their relations are fundamentally intertwined in the basic visual comprehension of videos. To reflect this in our benchmark, we have constructed an object-relation hallucination set. This set is divided into three distinct categories: subject, relation, and object. To elaborate, there is a wealth of prior research that has focused on annotating objects and relations within video content. In order to leverage this existing body of work and avoid duplicating efforts, we have developed our object-relation hallucination sets based on these pre-existing datasets. Specifically, we have utilized the VidOR~\citep{vidor1, vidor2} and VidVRD~\citep{vidvrd} datasets as the foundation for our set, ensuring that we build upon the annotations they provide without unnecessary redundancy in human annotation labor. For subject and object, we following the paradigm of POPE~\citep{POPE}, use the template ``Is there a/an \{object\} in the video?''. For relation, for spatial relation, we use the template ``Is \{subject\} \{relation\} \{object\} in the video?'', for temporal relation, or action, we use the template ``Does/Do \{subject\} \{relation\} \{object\} in the video?''. For hallucinated questions, instead of randomly selecting from existing object pools, we ask annotator to write semantic different but visually similar object, and anatomy of the relation word. After obtaining the question automatically, we manually check the question answer pairs. As a result, we get 200 basic-hallucination question-answer pairs, totally 400 question-answer pairs and 183 videos.

\paragraph{Temporal Hallucination}
Temporal information is the most fundamental difference between video input and image input. However, most recent video-text LLMs mainly focus on factual contents in short videos, ignoring the temporal information of multiple events in long videos. Recent studies \citep{singularity, nextgqa} found that they have almost no ability to perform time-related tasks such as video grounding.

The understanding of temporal information in videos contains two aspects: where in the video timeline is an event located, and how long it lasts. 
Therefore, we plan to test 3 time-related capabilities of video-text models: (1) 50 examples for \textbf{absolute temporal understanding}, where the model is asked to answer the position of an event in the video, such as whether it is located at the beginning or at the end of the video, (2) 75 examples for \textbf{relative temporal understanding}, where the model is asked to determine the order of two events, and (3) 75 examples for \textbf{event length understanding}, where the models is asked to compare the duration length of two events. 

We construct the temporal hallucination benchmark based on the val set of ActivityNet Captions ~\citep{anet}, a popular temporal video grounding dataset with event-level video annotations: each event is labeled with a text description and a time span (start sec, end sec) in the video. 
For absolute temporal understanding examples, we sampled 50 events from 50 different videos that are entirely located in the first or last third of its video. For events in the first third of its video, we use ``Does \{event\} happen at the beginning of the video?'' as the basic question, and ``Does \{event\} happen at the end of the video?'' as the hallucinated question, and vise versa for the events in the last thirds of its video.
For relative temporal understanding examples, we sampled 75 pairs of events from 75 different videos which satisfies that the end time of the first event is at least $0.1t$ earlier than the start time of the second event, where $t$ is the duration of the video. We randomly choose ``Does \{event1\} happen earlier than \{event2\}?'' or ``Does \{event2\} happen later than \{event1\}?'' as the basic question, and ``Does \{event1\} happen later than \{event2\}?'' or ``Does \{event2\} happen earlier than \{event1\}?'' as the hallucinated question.
For event length understanding examples, we also sampled 75 pairs of events from 75 different videos which satisfies that the duration of the first event is at least 2 times longer than the duration of the second event. We randomly choose ``Is the event \{event1\} longer than \{event2\}?'' or ``Is the event \{event2\} shorter than \{event1\}?'' as the basic question, and ``Is the event \{event1\} shorter than \{event2\}?'' or ``Is the event \{event2\} longer than \{event1\}?'' as the hallucinating question. As a result, we get 200 basic-hallucination question-answer pairs, totally 400 question-answer pairs and 165 videos.

\paragraph{Semantic Detail Hallucination}
Recent studies increasingly emphasize the importance of object attributes in visual analysis. A number of these works~\citep{mmhal-bench, amber, mitigating, fohe, correlationqa} delve deeper, attempting to uncover a broader spectrum of visual details such as optical character recognition (OCR), object counting, and environmental context. Given the challenge of creating a definitive classification for the diverse range of visual details, our work consolidates these attributes under the umbrella term ``other semantic details.'' This encompasses OCR, attribute recognition, viewpoint identification, and scene understanding, among others. It has been observed that most vision-language models exhibit deficiencies in accurately identifying these nuanced details. To address this gap, we have specifically developed a dataset aimed at enhancing the detection of hallucinations related to these other semantic details.

To enhance our understanding of the detailed semantic content in videos, we have adopted a technique inspired by contrastive learning, which involves identifying distinctions by drawing comparisons. In particular, we utilize video clips from the HawkEye dataset~\citep{hawkeye}, which segments lengthy videos into shorter clips based on PySceneDetect, and then merging clips that share high semantic similarity. These clips are designed to encapsulate distinct semantic meanings. Once we have these video clips, we compute the CLIP score among them to measure their semantic relatedness. We then select pairs of video clips with a CLIP score exceeding $0.85$ for further analysis. Subsequently, we engage annotators to pinpoint the semantic disparities between these selected pairs of clips. Based on these differences, the annotators are tasked with creating a set of basic-hallucinated question-answer pairs. Throughout this process, we instruct the annotators to pay particular attention to various aspects of the semantic details within the clips. As a result, we get 200 basic-hallucination question-answer pairs, totally 400 question-answer pairs and 400 videos.

\paragraph{Extrinsic Factual Hallucination}
Extrinsic factual hallucinations, which are unverifiable against the original source material yet remain factually consistent, may sometimes be considered beneficial. Certain studies suggest that these hallucinations could enhance the richness and informativeness of the generated content by providing additional background knowledge. However, the usefulness of such hallucinations is highly context-dependent. In the field of text summarization, any form of hallucination is undesirable, as accuracy and fidelity to the source are paramount. Conversely, in conversational agents, there is an expectation for the generation of varied and imaginative responses, where factual hallucinations might be more acceptable. Therefore, in our research, we do not seek to determine the overall value of factual hallucinations. Instead, our focus is on developing models that can detect and identify these hallucinations.

At \bench, we meticulously curate two distinct categories of videos: instructional videos and course lectures. Our selection criteria prioritize content that is rich in knowledge and can be understood independently of extrinsic information, thereby minimizing potential ambiguity. For the instructional category, we extract samples from the YouCook~\citep{youcook} and COIN~\citep{coin} datasets. These include cooking recipes and guides for various activities. In the realm of course lectures, our focus is on the EDUVSUM dataset~\citep{eduvsum}, which features online courses in fields like computer science, the history of science, and engineering. To simplify matters, we intentionally choose shorter videos from these sources. When dealing with instructional videos, our process begins by identifying the steps outlined in each tutorial. To formulate basic questions, we adopt a template that reads: ``Based on the video, should we \{step\} when we \{target\}?'' For hallucinated questions, we pay special attention to the final steps, as they typically contain detailed information and dropping these steps won't make the instructional video incomplete.  We select these as the focal point for \bench. Specifically, we trim videos and remove the content related the these stepsand pose questions in the same format.

For course lectures, we first instruct annotators to condense the content into a summary that captures the key points. We then generate basic questions following the pattern: ``Does the video provide information that allows us to learn that \{summary\}?'' For hallucinated questions, we challenge annotators to alter the summary or incorporate knowledge not covered in the video. This ensures that while the summary remains factual, it cannot be solely derived from the video content. Moreover, we ask the annotator to write an explanation for these hallucinated questions. The explanation point out the reason and the location of the hallucinated question.  Through this meticulous process, we have compiled a dataset of 200 paired basic and hallucination questions and answers, resulting in a total of 400 question-answer pairs and 200 videos, where 150 videos are instructional videos and 50 videos are online courses.

\paragraph{Extrinsic Non-facutal Hallucination}
Hallucinations that are not based on factual information can be particularly harmful due to their distorted content. Recently, there has been a growing interest among researchers in addressing this issue. At \bench, we have specifically developed a dataset to aid in the detection of non-factual hallucinations.

To ensure consistency with the detection of factual hallucinations, we have selected the same set of videos and corresponding basic questions used in the factual hallucination dataset. For the creation of hallucinated content, we instructed annotators to manually alter the basic questions or infuse them with counterfactual information. The format of these questions remains identical to that used in the factual context. As a result of this process, we have compiled a comprehensive dataset consisting of 400 question-answer pairs linked to 200 videos.

\subsection{Implementation Details}
\label{supp:implement}

\subsubsection{Setups for baselines}
\label{supp:implement-setup}
In our experiment, we choose the 7B-level video language models for fair comparison, including VideoChatGPT~\citep{videochatgpt}, Valley2~\citep{vally}, Video-LLaMA2~\citep{videollama}, VideoChat2~\citep{videochat}, VideoLLaVA~\citep{videollava}, LLaMA-VID~\citep{llamavid}, VideoLaVIT~\citep{videolavit}, and MiniGPT4-Video~\citep{minigpt4video}. To assure a fair comparison, we take the default hyper-parameter of these models, including ``max\_new\_tokens'', ``do\_sample'', ``temperature'', ``num\_beams'', and ``num\_of\_frames''.  To further reveal the potential of scaling law, we additionally add the existing best-performed closed-source model Gemini-1.5-Pro~\citep{gemini} as the current upbound of existing LVLMs. For the Gemini-1.5-Pro, we take the fps as 1, and to improve the evaluation efficiency, we set the max number of frames as 128. 

\subsubsection{Prompt for the Overall Results}
\label{supp:implement-eval}
To evaluate responses, we append the prompt ``Answer the question using 'yes' or 'no'.'' to the end of each question. We then compare the answer to ``yes'' or ``no'' to calculate accuracy.

\subsubsection{Implementation Detail of the Self-PEP}
\label{supp:implement-selfpep}

\begin{figure}
    \centering
    \includegraphics[width=\linewidth]{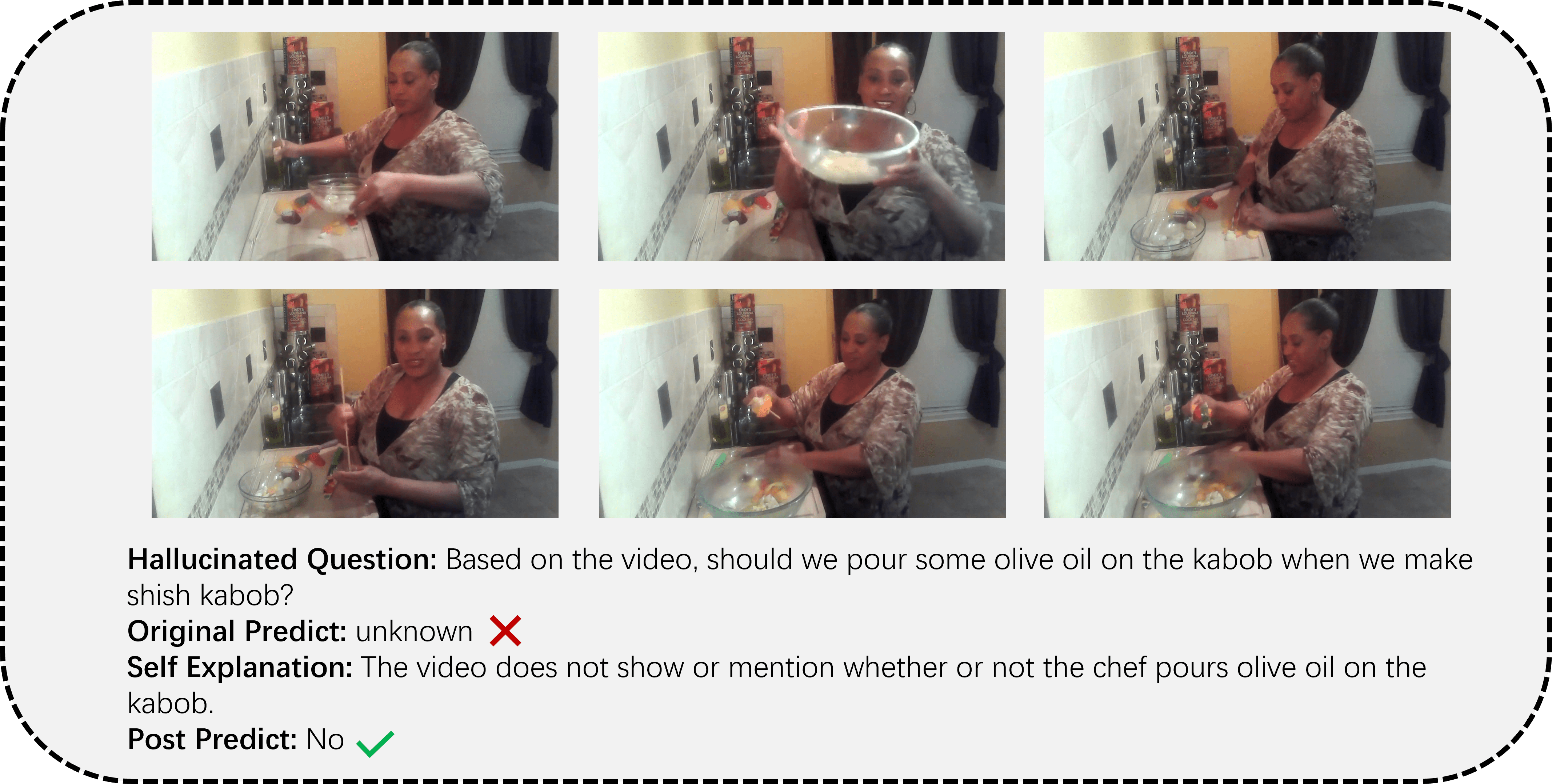}
    \caption{Qualitative Analysis: In the video observed, the woman initially drizzles olive oil solely on the shrimp, neglecting to do the same for the kabobs. A self-generated explanation can accurately identify the rationale behind this action and provide a clarified response.}
    \label{fig:selfpep-quality}
\end{figure}

As illustrated in ~\cref{fig:self-pep}, there are two main components of the Self-PEP: the self-improvement and the self-explain. For the self-improvement, we ask the model to extract the visual knowledge~\citep{woodpecker}. In our implementation, we ask the model to generate the caption for simplicity, the prompt is ``Describe the video: ''. After obtaining the caption, we ask the model to predict the answer the question with the input of both the video and the self-generated caption. The prompt is ``Description: \{description\} Please provide a clear response to the question below by watching the video. If necessary, you can also use the accompanying Description to help refine your answer. Your response should be a simple 'yes' or 'no'. Question: \{question\} Answer the question using 'yes' or 'no': ''. We further apply the self-explain to the model, by asking the model to explain-then-predict to get the final answer.  The prompt is ``Description: \{description\} Please offer a detailed explanation for your answer to the following question. After explaining, verify the accuracy of the information you've used in your explanation. Once you've confirmed the facts, please respond to the question with a simple 'yes' or 'no'. Question: \{question\} Answer: \{predict\} Answer the question using 'yes' or 'no': ''. Furthermore, we illustrate the qualitative study of the Self-PEP framework on Gemini-1.5-Pro in ~\cref{fig:selfpep-quality}, we find that with the explanation, the model is able to correct its answer.

\subsection{Explanation Analysis}
\label{supp:exp-analysis}

\begin{figure}[h]
    \centering
    \includegraphics[width=\linewidth]{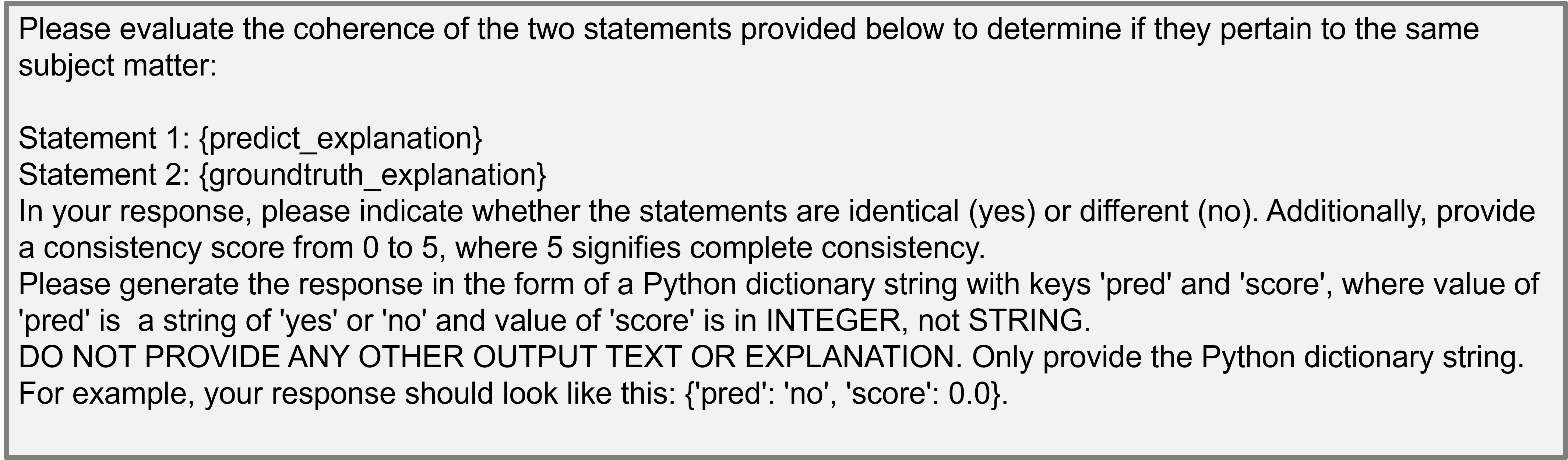}
    \caption{Prompt for Analysis of the Explanations}
    \label{fig:analysis}
\end{figure}

Upon comparing \cref{fig:PEP_res} with \cref{fig:fact-det}, it becomes evident that there is a correlation between the model's self-explanation capability and its ability to detect factual information. To further explore this relationship, we evaluate how well the self-generated explanations align with the established ground truth explanations. For this purpose, we employ GPT-4 to conduct a consistency analysis, utilizing the prompt outlined in \cref{fig:analysis}. The results, presented in \cref{tab:coherence}, indicate that the quality of the explanations has a significant impact on the model's proficiency in identifying instances of hallucination. Consequently, our goal is to enhance the model's capacity to produce accurate and reliable explanations as a means to mitigate the issue of hallucination.

\begin{table}[h]
    
    \caption{\textbf{Coherence of the self-generated explanation and the annotated explanation}}
    \label{tab:coherence}
    \centering
    \begin{tabular}{l|c|c}
        \toprule
        \textbf{Models} & \textbf{Consistent Rate} & \textbf{GPT4 Score} \\\midrule
         VideoChatGPT~\citep{videochatgpt} &  33.5 & 2.555 \\ 
         Valley2~\citep{vally} & 39 & 2.34 \\
         Video-LLaMA-2~\citep{videollama} & 8.5 & 1.175 \\
         VideoChat2~\citep{videochat} & 38 & 2.42 \\
         VideoLLaVA~\citep{videollava} &  36.5 & 2.45 \\ 
         LLaMA-VID~\citep{llamavid} &  42 & 2.625 \\ 
         VideoLaVIT~\citep{videolavit} &  36 & 2.435 \\
         MiniGPT4-Video~\citep{minigpt4video} & 31.5 & 2.085 \\
         Gemini-1.5-Pro~\citep{gemini} & 37 & 2.54 \\
         \bottomrule
    \end{tabular}
\end{table}

\subsection{Detailed Quantitative Results}
\label{supp:detail-quant}
In this section, we provide the detailed quantitative results on \bench, including the overall results (\cref{supp:overall}) and the results with Self-PEP (\cref{supp:self-pep}).

\subsubsection{The Detailed Results on \bench}
\label{supp:overall}

In this section, we present the comprehensive results of various LVLMs across different settings of \bench. \cref{tab:overall-obj-rel} details the outcomes for object-relation hallucination; \cref{tab:overall-temporal} displays the findings for temporal hallucination; \cref{tab:overall-semantic} outlines the results for semantic detail hallucination; \cref{tab:overall-fact} depicts the results for extrinsic factual hallucination; and \cref{tab:overall-nonfact} provides the results for extrinsic non-factual hallucination.

\begin{table}[h]
    
    \caption{\textbf{Overall Results on Object Relation Hallucination}}
    \label{tab:overall-obj-rel}
    \centering
    \resizebox{\linewidth}{!}{
    \begin{tabular}{lcccc}
        \toprule
        \textbf{Models}  & \textbf{Basic Question} & \textbf{Hallucinated Question} & \textbf{Overall Accuracy}  \\\midrule
        VideoChatGPT~\citep{videochatgpt} & 95.5 & 7 & 6 \\
         Valley2~\citep{vally}& 80.5 & 10.5 & 4.5  \\ 
         Video-LLaMA2~\citep{videollama}& 88.5 & 21.5 & 18 \\
         VideoChat2~\citep{videochat}  &26 & 41.5 & 10.5 \\
         VideoLLaVA~\citep{videollava}  &  95 & 38 & 34.5 \\
         LLaMA-VID~\citep{llamavid}  &  78.5 & 59 & 43.5 \\
         LLaMA-VID-13B~\citep{llamavid}  &  87.5 & 55.5 & 44.5 \\
         VideoLaVIT~\citep{videolavit}  & 94.5 & 39 & 35.5 \\
         MiniGPT4-Video~\citep{minigpt4video}  & 80.5 & 34.5 & 27.5 \\
         PLLaVA~\citep{xu2024pllava}  & 76 & 76.5 & 60 \\
         LLaVA-NeXT-Video-DPO~\citep{zhang2024llavanextvideo}  & 72 & 73 & 51.5 \\ 
         Gemini-1.5-Pro~\citep{gemini}  &  84.5 & 56 & 52 \\ 
         \bottomrule
    \end{tabular}}
\end{table}

\begin{table}[h]
    
    \caption{\textbf{Overall Results on Temporal Hallucination}}
    \label{tab:overall-temporal}
    \centering
    \resizebox{\linewidth}{!}{
    \begin{tabular}{lcccc}
        \toprule
        \textbf{Models}  & \textbf{Basic Question} & \textbf{Hallucinated Question} & \textbf{Overall Accuracy}  \\\midrule
        VideoChatGPT~\citep{videochatgpt} & 100 & 0 & 0 \\
         Valley2~\citep{vally}& 25 & 11.5 & 3  \\ 
         Video-LLaMA2~\citep{videollama}& 91.5 & 8.5 & 7.5 \\
         VideoChat2~\citep{videochat}  &23.5 & 25 & 7.5 \\
         VideoLLaVA~\citep{videollava}  &  97.5 & 13.5 & 13.5 \\
         LLaMA-VID~\citep{llamavid}  &  86 & 25 & 21 \\
         LLaMA-VID-13B~\citep{llamavid}  &  78.5 & 35 & 27 \\
         VideoLaVIT~\citep{videolavit}  & 88.5 & 27 &  25.5 \\
         MiniGPT4-Video~\citep{minigpt4video}  & 68.5 & 27 & 18 \\ 
         PLLaVA~\citep{xu2024pllava}  & 46.5 & 58 & 23.5 \\
         LLaVA-NeXT-Video-DPO~\citep{zhang2024llavanextvideo}  & 53 & 61 & 28 \\ 
         Gemini-1.5-Pro~\citep{gemini}  &  81.5 & 19 & 18.5 \\ 
         \bottomrule
    \end{tabular}}
\end{table}

\begin{table}[h]
    
    \caption{\textbf{Overall Results on Semantic Detail Hallucination}}
    \label{tab:overall-semantic}
    \centering
    \resizebox{\linewidth}{!}{
    \begin{tabular}{lcccc}
        \toprule
        \textbf{Models}  & \textbf{Basic Question} & \textbf{Hallucinated Question} & \textbf{Overall Accuracy}  \\\midrule
        VideoChatGPT~\citep{videochatgpt} &96.5 & 4 & 2 \\
         Valley2~\citep{vally}& 86.5 & 6.5 & 2.5  \\ 
         Video-LLaMA2~\citep{videollama}& 99 & 1.5 & 1 \\
         VideoChat2~\citep{videochat}  &33 & 26 & 9 \\
         VideoLLaVA~\citep{videollava}  &  97 & 14 & 12 \\
         LLaMA-VID~\citep{llamavid}  &  89 & 24 & 17 \\
         LLaMA-VID-13B~\citep{llamavid}  &  90.5 & 30 & 25.5 \\
         VideoLaVIT~\citep{videolavit}  & 96.5 & 13 & 10.5 \\
         MiniGPT4-Video~\citep{minigpt4video}  & 78.5 & 27.5 & 23.5 \\ 
         PLLaVA~\citep{xu2024pllava}  & 83 & 71.5 & 57 \\
         LLaVA-NeXT-Video-DPO~\citep{zhang2024llavanextvideo}  & 63.5 & 69 & 38 \\ 
         Gemini-1.5-Pro~\citep{gemini}  &  89 & 63 & 53.5 \\ 
         \bottomrule
    \end{tabular}}
\end{table}

\begin{table}[h]
    
    \caption{\textbf{Overall Results on extrinsic Factual Hallucination}}
    \label{tab:overall-fact}
    \centering
    \resizebox{\linewidth}{!}{
    \begin{tabular}{lcccc}
        \toprule
        \textbf{Models}  & \textbf{Basic Question} & \textbf{Hallucinated Question} & \textbf{Overall Accuracy}  \\\midrule
        VideoChatGPT~\citep{videochatgpt} & 86.5 & 13.5 & 7 \\
         Valley2~\citep{vally}& 14 & 15 & 0.5  \\ 
         Video-LLaMA2~\citep{videollama}& 88 & 8.5 & 6.5 \\
         VideoChat2~\citep{videochat}  &32 & 16.5 & 7 \\
         VideoLLaVA~\citep{videollava}  &  93 & 4.5 & 3 \\
         LLaMA-VID~\citep{llamavid}  &  98 & 2.5 & 2.5 \\
         LLaMA-VID-13B~\citep{llamavid}  &  85 & 17.5 & 12.5 \\
         VideoLaVIT~\citep{videolavit}  & 97.5 & 6& 4 \\
         MiniGPT4-Video~\citep{minigpt4video}  & 86 & 16.5 & 12 \\ 
         PLLaVA~\citep{xu2024pllava}  & 85 & 18 & 9.5 \\
         LLaVA-NeXT-Video-DPO~\citep{zhang2024llavanextvideo}  & 62.5 & 41 & 14 \\ 
         Gemini-1.5-Pro~\citep{gemini}  &  82 & 19 & 16.5 \\ 
         \bottomrule
    \end{tabular}}
\end{table}

\begin{table}[h]
    
    \caption{\textbf{Overall Results on extrinsic Non-factual Hallucination}}
    \label{tab:overall-nonfact}
    \centering
    \resizebox{\linewidth}{!}{
    \begin{tabular}{lcccc}
        \toprule
        \textbf{Models}  & \textbf{Basic Question} & \textbf{Hallucinated Question} & \textbf{Overall Accuracy}  \\\midrule
        VideoChatGPT~\citep{videochatgpt} & 85.5 & 27.5 & 17 \\
         Valley2~\citep{vally}& 16 & 14 & 3.5  \\ 
         Video-LLaMA2~\citep{videollama}& 87.5 & 23.5 & 17 \\
         VideoChat2~\citep{videochat}  &34 & 20 & 0.5 \\
         VideoLLaVA~\citep{videollava}  &  93 & 31.5 & 26 \\
         LLaMA-VID~\citep{llamavid}  &  98 & 22.5 & 21 \\
         LLaMA-VID-13B~\citep{llamavid}  &  84.5 & 46.5 & 36.5 \\
         VideoLaVIT~\citep{videolavit}  & 97.5 & 21.5 & 19 \\
         MiniGPT4-Video~\citep{minigpt4video}  & 83.5 & 37.5 & 30.5 \\ 
         PLLaVA~\citep{xu2024pllava}  & 85 & 53.5 & 40.5 \\
         LLaVA-NeXT-Video-DPO~\citep{zhang2024llavanextvideo}  & 61.5 & 60.5 & 28.5 \\ 
         Gemini-1.5-Pro~\citep{gemini}  &  81 & 54.5 & 48.5 \\ 
         \bottomrule
    \end{tabular}}
\end{table}

\subsubsection{The Detailed Results of Self-PEP}
\label{supp:self-pep}

In this section, we further show the detailed results of Self-PEP on different settings of \bench. Table \cref{tab:selfpep-obj-rel} details the outcomes for object-relation hallucination; Table \cref{tab:selfpep-temporal} displays the findings for temporal hallucination; Table \cref{tab:selfpep-semantic} outlines the results for semantic detail hallucination; Table \cref{tab:selfpep-fact} depicts the results for extrinsic factual hallucination; and Table \cref{tab:selfpep-nonfact} provides the results for extrinsic non-factual hallucination.

\begin{table}[h!]
    
    \caption{\textbf{Self-PEP Results on Object Relation Hallucination}}
    \label{tab:selfpep-obj-rel}
    \centering
    \resizebox{\linewidth}{!}{
    \begin{tabular}{lcccc}
        \toprule
        \textbf{Models}  & \textbf{Basic Question} & \textbf{Hallucinated Question} & \textbf{Overall Accuracy}  \\\midrule
        VideoChatGPT~\citep{videochatgpt} & 95.5 & 7 & 6 \\
        \rowcolor{lightgray}
        +Self-PEP &72.5 & 53 & 33.5 \\
         Valley2~\citep{vally}& 80.5 & 10.5 & 4.5  \\  
         \rowcolor{lightgray}
        +Self-PEP &20.5 & 31 & 10 \\
         Video-LLaMA2~\citep{videollama}& 88.5 & 21.5 & 18 \\
         \rowcolor{lightgray}
        +Self-PEP &75 & 27.5 & 12 \\
         VideoChat2~\citep{videochat}  &26 & 41.5 & 10.5 \\
         \rowcolor{lightgray}
        +Self-PEP &61 & 58.5 & 34 \\
         VideoLLaVA~\citep{videollava}  &  95 & 38 & 34.5 \\
         \rowcolor{lightgray}
        +Self-PEP &72 & 75 & 52 \\
         LLaMA-VID~\citep{llamavid}  &  78.5 & 59 & 43.5 \\
         \rowcolor{lightgray}
        +Self-PEP &60 & 74.5 & 44.5 \\
         VideoLaVIT~\citep{videolavit}  & 94.5 & 39 & 35.5 \\
         \rowcolor{lightgray}
        +Self-PEP &90.5 & 29 & 24 \\
         MiniGPT4-Video~\citep{minigpt4video}  & 80.5 & 34.5 & 27.5 \\
         \rowcolor{lightgray}
        +Self-PEP &72 & 64.5 & 47 \\
        PLLaVA~\citep{xu2024pllava}  & 76 & 76.5 & 60 \\
        \rowcolor{lightgray}
        +Self-PEP &67.5 & 74 & 52 \\
         LLaVA-NeXT-Video-DPO~\citep{zhang2024llavanextvideo}  & 72 & 73 & 51.5 \\ 
         \rowcolor{lightgray}
        +Self-PEP &67 & 79.5 & 51 \\
         Gemini-1.5-Pro~\citep{gemini}  &  84.5 & 56 & 52 \\ 
         \rowcolor{lightgray}
        +Self-PEP & 62.5 & 84 & 56 \\
         \bottomrule
    \end{tabular}}
\end{table}

\begin{table}[h!]
    
    \caption{\textbf{Self-PEP Results on Temporal Hallucination}}
    \label{tab:selfpep-temporal}
    \centering
    \resizebox{\linewidth}{!}{
    \begin{tabular}{lcccc}
        \toprule
        \textbf{Models}  & \textbf{Basic Question} & \textbf{Hallucinated Question} & \textbf{Overall Accuracy}  \\\midrule
        VideoChatGPT~\citep{videochatgpt} & 100 & 0 & 0 \\
        \rowcolor{lightgray}
        +Self-PEP &95.5 & 8 & 4.5 \\
         Valley2~\citep{vally}& 25 & 11.5 & 3  \\
         \rowcolor{lightgray}
        +Self-PEP &21.5 & 15 & 5.5 \\
         Video-LLaMA2~\citep{videollama}& 91.5 & 8.5 & 7.5 \\
         \rowcolor{lightgray}
        +Self-PEP &57.5 & 23 & 5.5 \\
         VideoChat2~\citep{videochat}  &23.5 & 25 & 7.5 \\
         \rowcolor{lightgray}
        +Self-PEP &47 & 34 & 15 \\
         VideoLLaVA~\citep{videollava}  &  97.5 & 13.5 & 13.5 \\
         \rowcolor{lightgray}
        +Self-PEP &97.5 & 5.5 & 5.5 \\
         LLaMA-VID~\citep{llamavid}  & 86 & 25 & 21 \\
         \rowcolor{lightgray}
        +Self-PEP &87.5 & 19.5 & 14 \\
         VideoLaVIT~\citep{videolavit}  & 88.5 &  27 & 25.5  \\
         \rowcolor{lightgray}
        +Self-PEP &100 & 0.5 & 0.5 \\
         MiniGPT4-Video~\citep{minigpt4video}  & 68.5 & 27 & 18 \\ 
         \rowcolor{lightgray}
        +Self-PEP &82 & 26 & 21.5 \\
        PLLaVA~\citep{xu2024pllava}  & 46.5 & 58 & 23.5 \\
        \rowcolor{lightgray}
        +Self-PEP &67.5 & 34 & 8 \\
         LLaVA-NeXT-Video-DPO~\citep{zhang2024llavanextvideo}  & 53 & 61 & 28 \\ 
         \rowcolor{lightgray}
        +Self-PEP &52 & 54.5 & 15 \\
         Gemini-1.5-Pro~\citep{gemini}  &  81.5 & 19 & 18.5 \\ 
         \rowcolor{lightgray}
        +Self-PEP & 66 & 57 & 44 \\
         \bottomrule
    \end{tabular}}
\end{table}

\begin{table}[h!]
    
    \caption{\textbf{Self-PEP Results on Semantic Detail Hallucination}}
    \label{tab:selfpep-semantic}
    \centering
    \resizebox{\linewidth}{!}{
    \begin{tabular}{lcccc}
        \toprule
        \textbf{Models}  & \textbf{Basic Question} & \textbf{Hallucinated Question} & \textbf{Overall Accuracy}  \\\midrule
        VideoChatGPT~\citep{videochatgpt} &96.5 & 4 & 2 \\
        \rowcolor{lightgray}
        +Self-PEP &72 & 42 & 22.5 \\
         Valley2~\citep{vally}& 86.5 & 6.5 & 2.5  \\ 
         \rowcolor{lightgray}
        +Self-PEP &12.5 & 16.5 & 1.5 \\
         Video-LLaMA2~\citep{videollama}& 99 & 1.5 & 1 \\
         \rowcolor{lightgray}
        +Self-PEP &79.5 & 10.5 & 8 \\
         VideoChat2~\citep{videochat}  &33 & 26 & 9 \\
         \rowcolor{lightgray}
        +Self-PEP &75 & 39.5 & 27.5 \\
         VideoLLaVA~\citep{videollava}  &  97 & 14 & 12 \\
         \rowcolor{lightgray}
        +Self-PEP &74 & 58.5 & 36 \\
         LLaMA-VID~\citep{llamavid}  &  89 & 24 & 17 \\
         \rowcolor{lightgray}
        +Self-PEP &50 & 80 & 36.5 \\
         VideoLaVIT~\citep{videolavit}  & 96.5 & 13 & 10.5 \\
         \rowcolor{lightgray}
        +Self-PEP &92 & 27 & 22 \\
         MiniGPT4-Video~\citep{minigpt4video}  & 78.5 & 27.5 & 23.5 \\ 
         \rowcolor{lightgray}
        +Self-PEP &77.5 & 44.5 & 34.5 \\
        PLLaVA~\citep{xu2024pllava}  & 83 & 71.5 & 57 \\
        \rowcolor{lightgray}
        +Self-PEP &64 & 78 & 46.5 \\
         LLaVA-NeXT-Video-DPO~\citep{zhang2024llavanextvideo}  & 63.5 & 69 & 38 \\ 
         \rowcolor{lightgray}
        +Self-PEP &50 & 80 & 37.5 \\
         Gemini-1.5-Pro~\citep{gemini}  &  89 & 63 & 53.5 \\ 
         \rowcolor{lightgray}
        +Self-PEP & 72.5 & 88.5 & 64 \\
         \bottomrule
    \end{tabular}}
\end{table}

\begin{table}[h!]
    
    \caption{\textbf{Self-PEP Results on extrinsic Factual Hallucination}}
    \label{tab:selfpep-fact}
    \centering
    \resizebox{\linewidth}{!}{
    \begin{tabular}{lcccc}
        \toprule
        \textbf{Models}  & \textbf{Basic Question} & \textbf{Hallucinated Question} & \textbf{Overall Accuracy}  \\\midrule
        VideoChatGPT~\citep{videochatgpt} & 86.5 & 13.5 & 7 \\
        \rowcolor{lightgray}
        +Self-PEP &74 & 28 & 14 \\
         Valley2~\citep{vally}& 14 & 15 & 0.5  \\  
         \rowcolor{lightgray}
        +Self-PEP &11.5 & 15.5 & 1.5 \\
         Video-LLaMA2~\citep{videollama}& 88 & 8.5 & 6.5 \\
         \rowcolor{lightgray}
        +Self-PEP &58 & 37 & 15.5 \\
         VideoChat2~\citep{videochat}  &32 & 16.5 & 7 \\
         \rowcolor{lightgray}
        +Self-PEP &57.5 & 38.5 & 19.5 \\
         VideoLLaVA~\citep{videollava}  &  93 & 4.5 & 3 \\
         \rowcolor{lightgray}
        +Self-PEP &80.5 & 23 & 11 \\
         LLaMA-VID~\citep{llamavid}  &  98 & 2.5 & 2.5 \\
         \rowcolor{lightgray}
        +Self-PEP &65.5 & 44 & 22 \\
         VideoLaVIT~\citep{videolavit}  & 97.5 & 6& 4 \\
         \rowcolor{lightgray}
        +Self-PEP &94.5 & 6.5 & 6 \\
         MiniGPT4-Video~\citep{minigpt4video}  & 86 & 16.5 & 12 \\ 
         \rowcolor{lightgray}
        +Self-PEP &80 & 26.5 & 14 \\
        PLLaVA~\citep{xu2024pllava}  & 85 & 18 & 9.5 \\
        \rowcolor{lightgray}
        +Self-PEP &63.5 & 45 & 16.5 \\
         LLaVA-NeXT-Video-DPO~\citep{zhang2024llavanextvideo}  & 62.5 & 41 & 14 \\ 
         \rowcolor{lightgray}
        +Self-PEP &28.5 & 72 & 11 \\
         Gemini-1.5-Pro~\citep{gemini}  &  82 & 19 & 16.5 \\ 
         \rowcolor{lightgray}
        +Self-PEP & 61.5 & 55 & 33 \\
         \bottomrule
    \end{tabular}}
\end{table}

\begin{table}[h!]
    
    \caption{\textbf{Self-PEP Results on extrinsic Non-factual Hallucination}}
    \label{tab:selfpep-nonfact}
    \centering
    \resizebox{\linewidth}{!}{
    \begin{tabular}{lcccc}
        \toprule
        \textbf{Models}  & \textbf{Basic Question} & \textbf{Hallucinated Question} & \textbf{Overall Accuracy}  \\\midrule
        VideoChatGPT~\citep{videochatgpt} & 85.5 & 27.5 & 17 \\
        \rowcolor{lightgray}
        +Self-PEP &75 & 49 & 30 \\
         Valley2~\citep{vally}& 16 & 14 & 3.5  \\
         \rowcolor{lightgray}
        +Self-PEP &10.5 & 14.5 & 4.5 \\
         Video-LLaMA2~\citep{videollama}& 87.5 & 23.5 & 17 \\
         \rowcolor{lightgray}
        +Self-PEP &65 & 52.6& 26 \\
         VideoChat2~\citep{videochat}  &34 & 20 & 0.5 \\
         \rowcolor{lightgray}
        +Self-PEP &47 & 50.5 & 21.5 \\
         VideoLLaVA~\citep{videollava}  &  93 & 31.5 & 26 \\
         \rowcolor{lightgray}
        +Self-PEP &80.5 & 53 & 34 \\
         LLaMA-VID~\citep{llamavid}  &  98 & 22.5 & 21 \\
         \rowcolor{lightgray}
        +Self-PEP &62.5 & 65 & 33 \\
         VideoLaVIT~\citep{videolavit}  & 97.5 & 21.5 & 19 \\
         \rowcolor{lightgray}
        +Self-PEP &94.5 & 15 & 13.5 \\
         MiniGPT4-Video~\citep{minigpt4video}  & 83.5 & 37.5 & 30.5 \\
         \rowcolor{lightgray}
        +Self-PEP &80 & 48.5 & 35.5 \\
        PLLaVA~\citep{xu2024pllava}  & 85 & 53.5 & 40.5 \\
        \rowcolor{lightgray}
        +Self-PEP &63.5 & 72 & 39.5 \\
         LLaVA-NeXT-Video-DPO~\citep{zhang2024llavanextvideo}  & 61.5 & 60.5 & 28.5 \\
         \rowcolor{lightgray}
        +Self-PEP &31.5 & 82.5 & 16 \\
         Gemini-1.5-Pro~\citep{gemini}  &  81 & 54.5 & 48.5 \\ 
         \rowcolor{lightgray}
        +Self-PEP & 67.5 & 92 & 63 \\
         \bottomrule
    \end{tabular}}
\end{table}

\clearpage

\subsection{Example Questions of \bench}
In this section, we provide more cases from \bench.
\label{supp:example}

\begin{figure}[h]
    \centering
    \includegraphics[width=\linewidth]{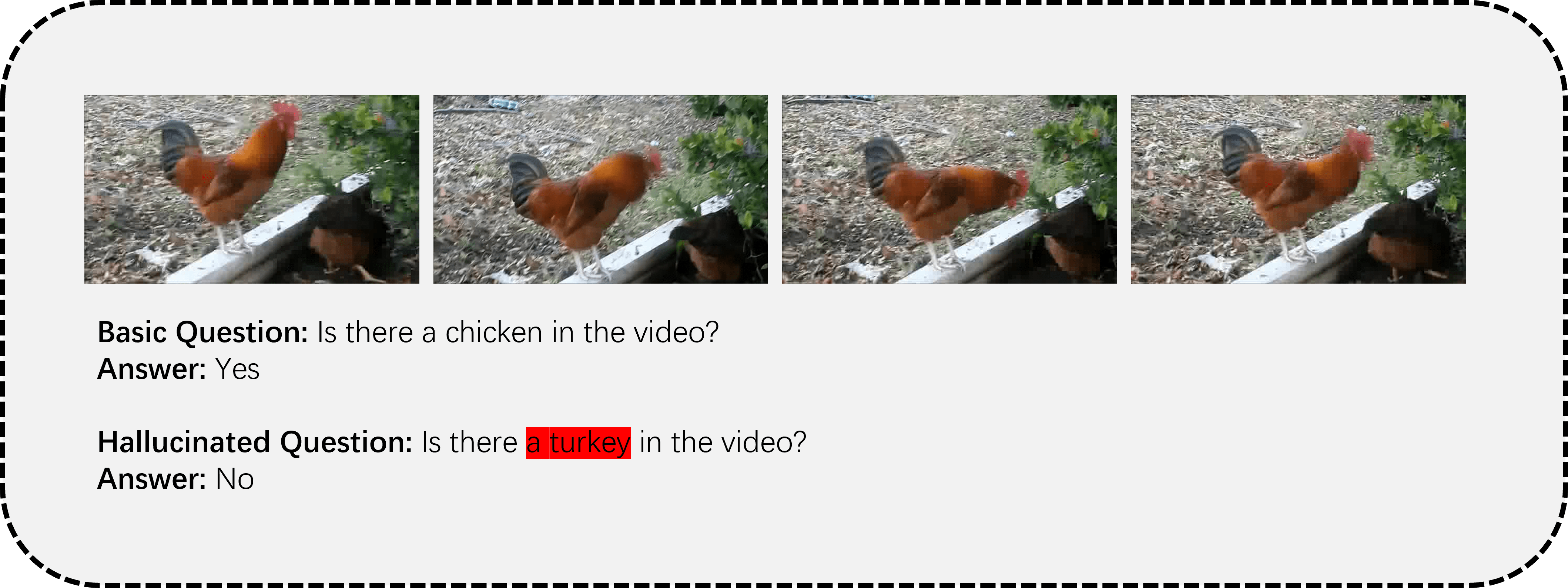}
    \caption{An example Basic-Hallucinated question pair of Object Hallucination}
    \label{fig:case-obj}
\end{figure}

\begin{figure}[h]
    \centering
    \includegraphics[width=\linewidth]{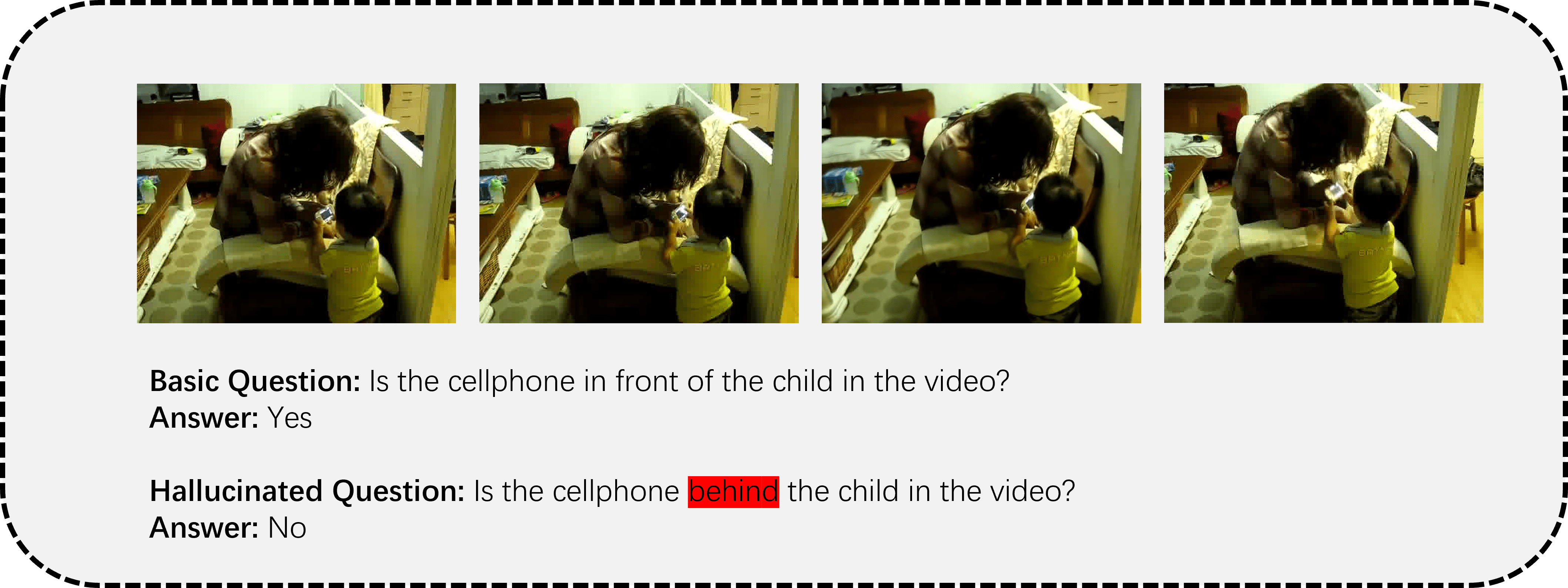}
    \caption{An example Basic-Hallucinated question pair of Spatial Relation Hallucination}
    \label{fig:case-rel_spatial}
\end{figure}

\begin{figure}[h]
    \centering
    \includegraphics[width=\linewidth]{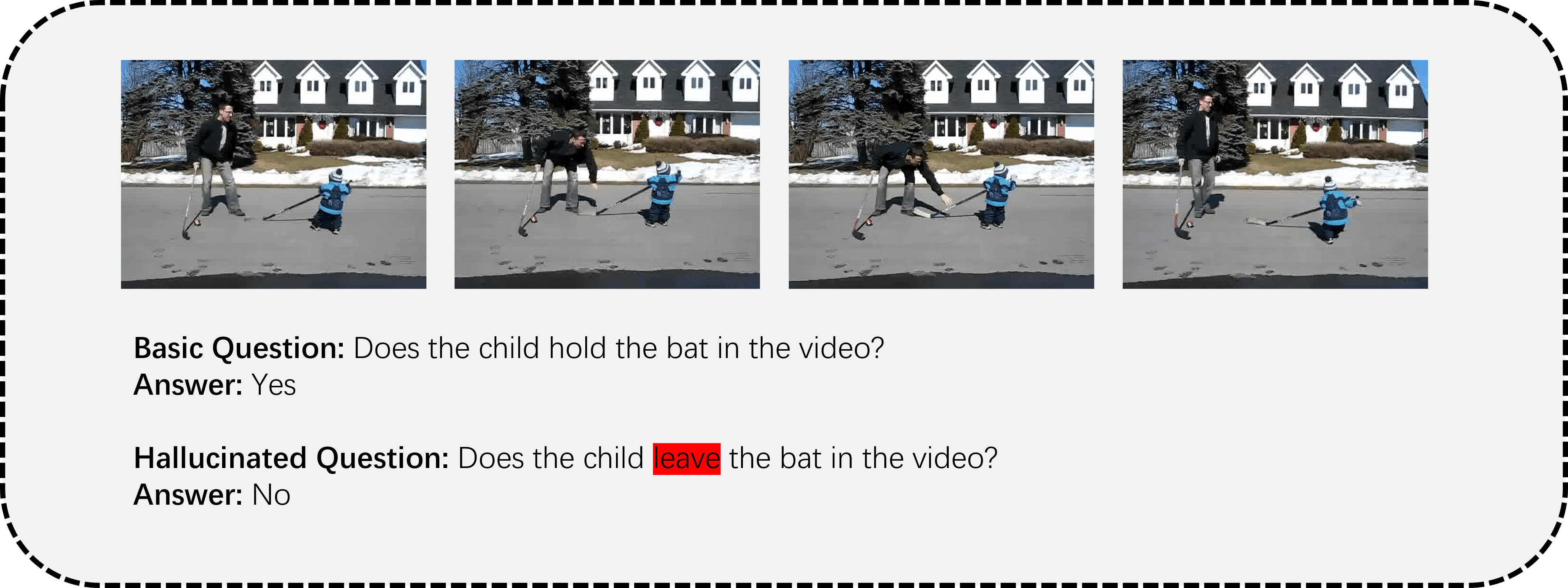}
    \caption{An example Basic-Hallucinated question pair of Temporal Relation Hallucination}
    \label{fig:case-rel_temp}
\end{figure}

\begin{figure}[h]
    \centering
    \includegraphics[width=\linewidth]{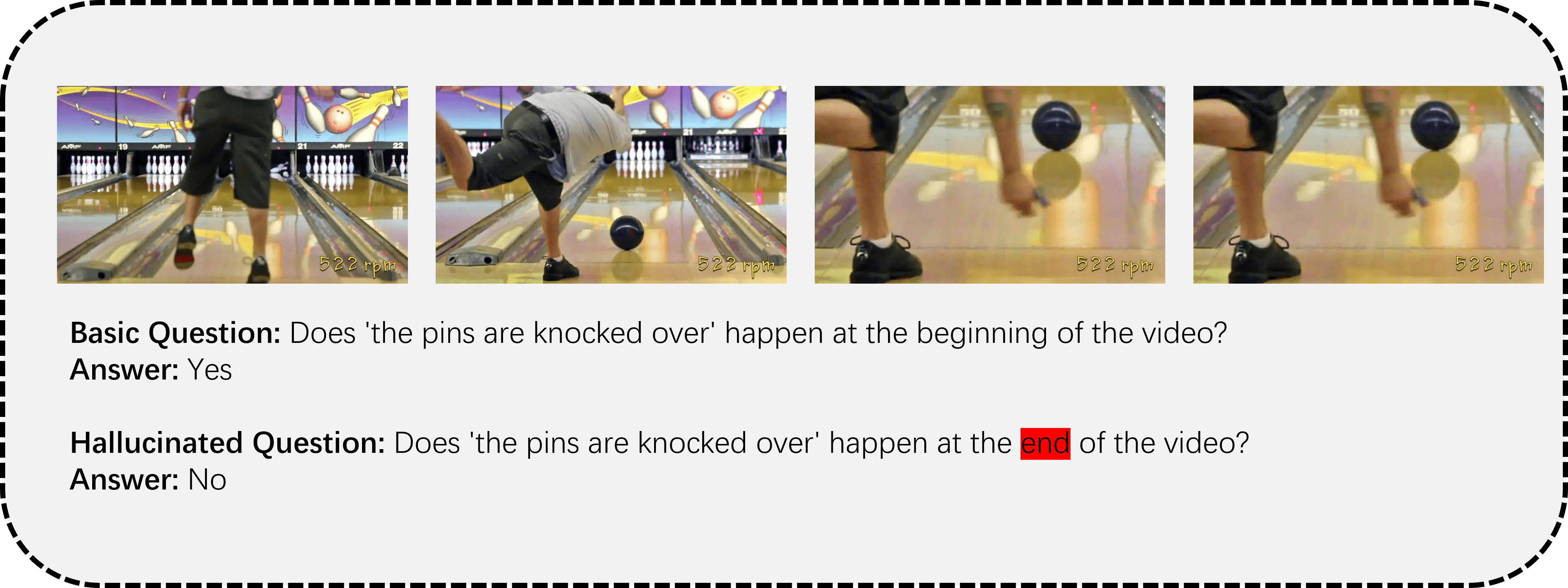}
    \caption{An example Basic-Hallucinated question pair of Absolute Temporal Hallucination}
    \label{fig:case-temporal_abl}
\end{figure}

\begin{figure}[h]
    \centering
    \includegraphics[width=\linewidth]{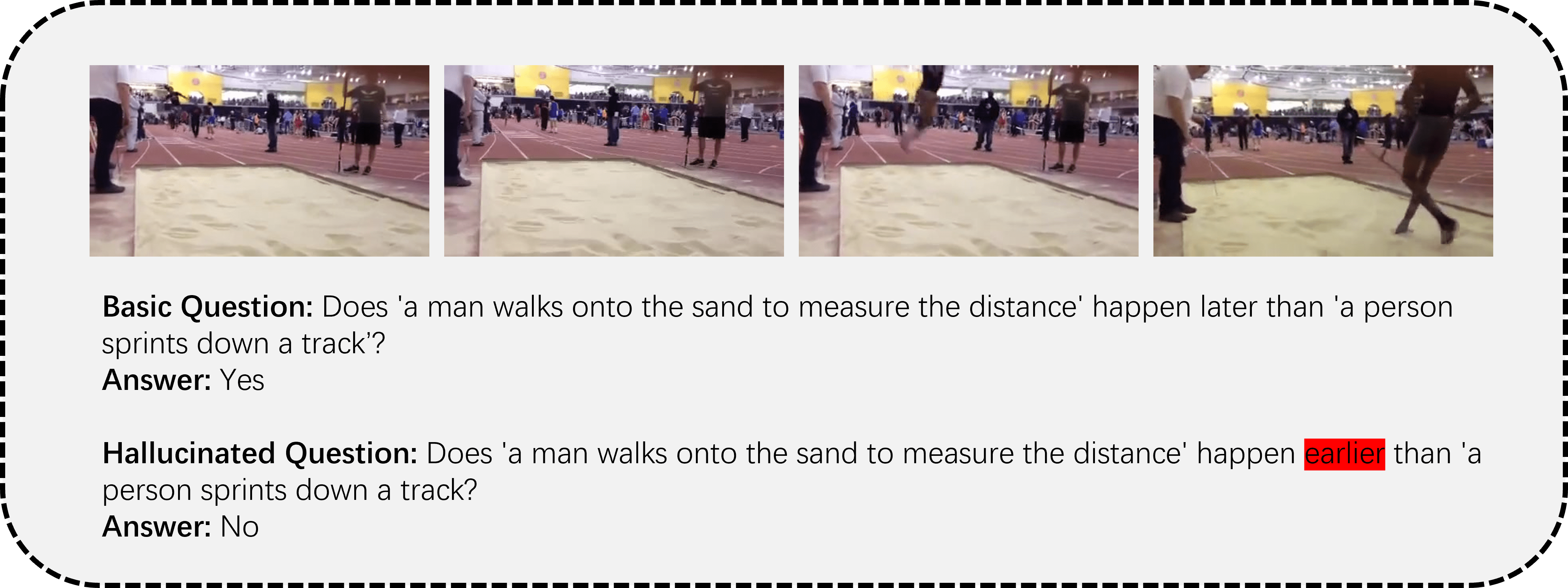}
    \caption{An example Basic-Hallucinated question pair of Relative Temporal Hallucination}
    \label{fig:case-temporal_rel}
\end{figure}

\begin{figure}[h]
    \centering
    \includegraphics[width=\linewidth]{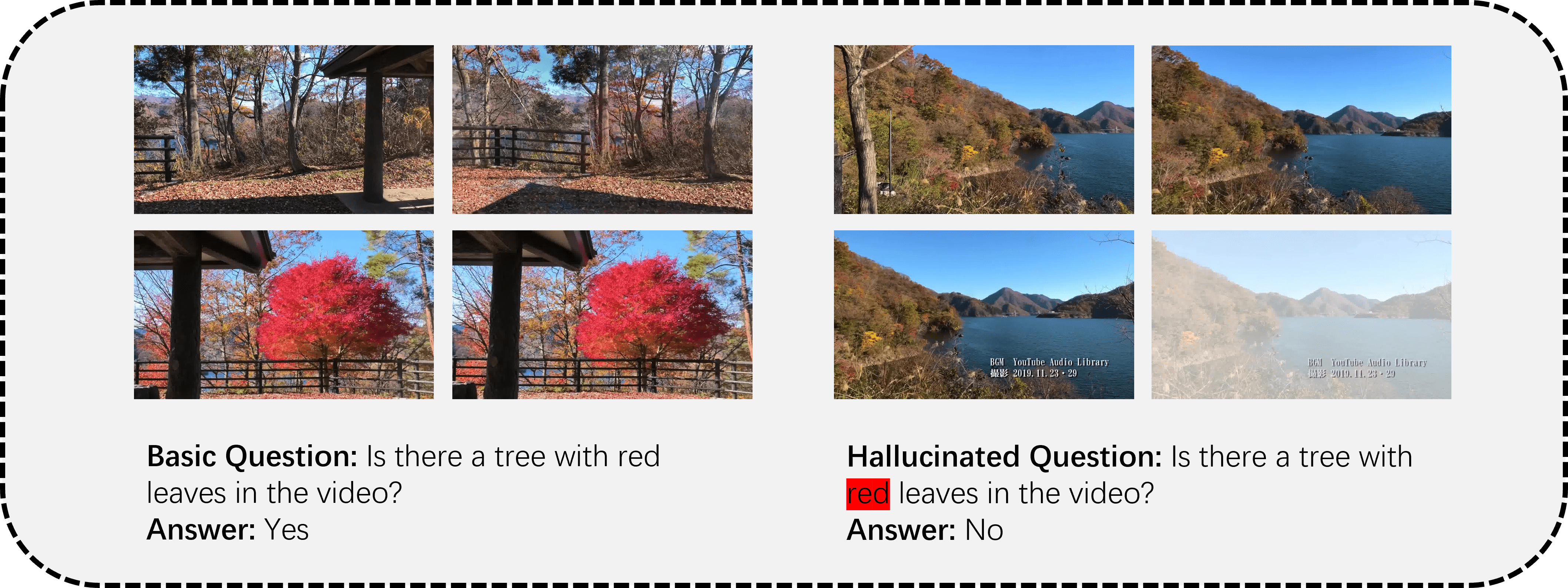}
    \caption{An example Basic-Hallucinated question pair of Semantic Detail (Attribution) Hallucination}
    \label{fig:case-semantic_attr}
\end{figure}

\begin{figure}[h]
    \centering
    \includegraphics[width=\linewidth]{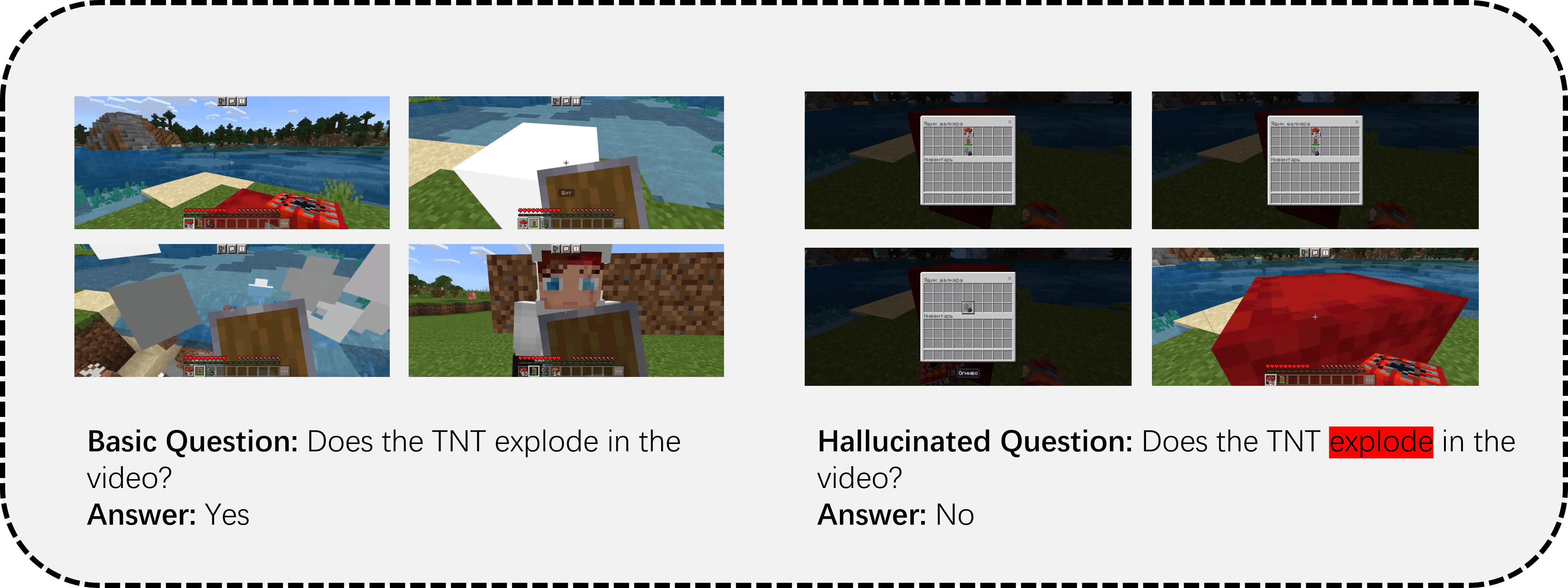}
    \caption{An example Basic-Hallucinated question pair of Semantic Detail (Event) Hallucination}
    \label{fig:case-semantic_event}
\end{figure}

\begin{figure}[h]
    \centering
    \includegraphics[width=\linewidth]{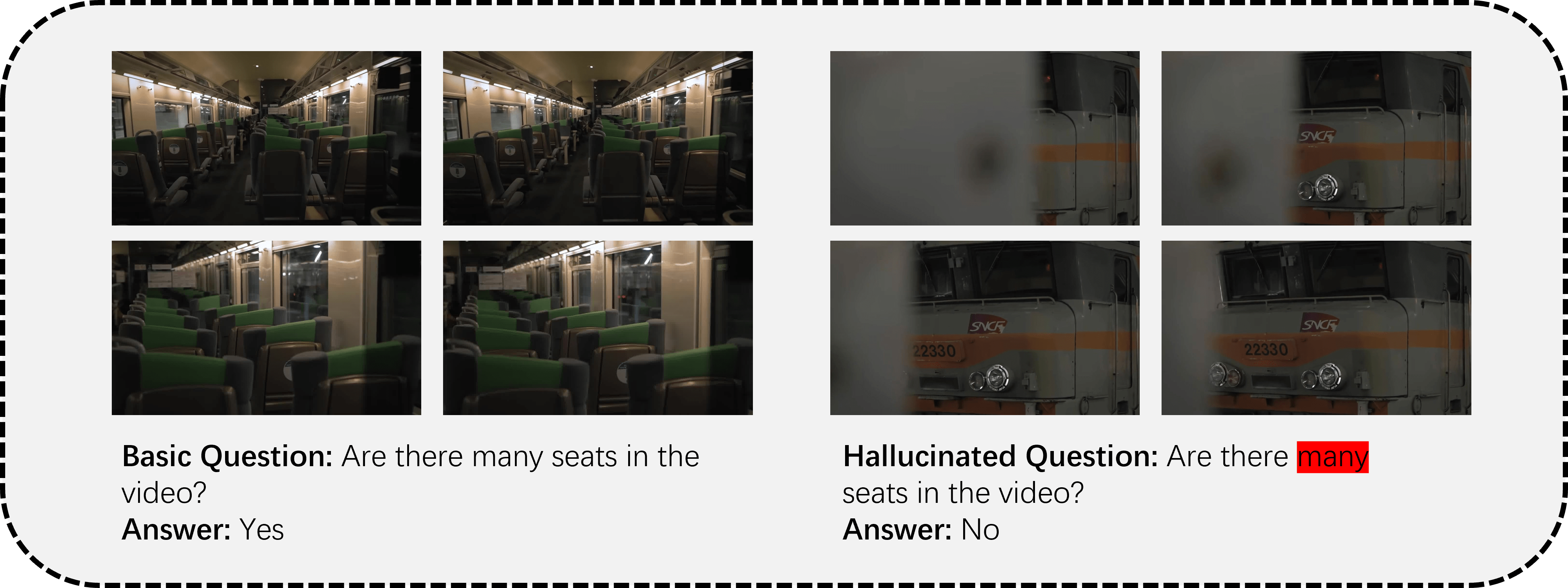}
    \caption{An example Basic-Hallucinated question pair of Semantic Detail (Count) Hallucination}
    \label{fig:case-semantic_count}
\end{figure}

\begin{figure}[h]
    \centering
    \includegraphics[width=\linewidth]{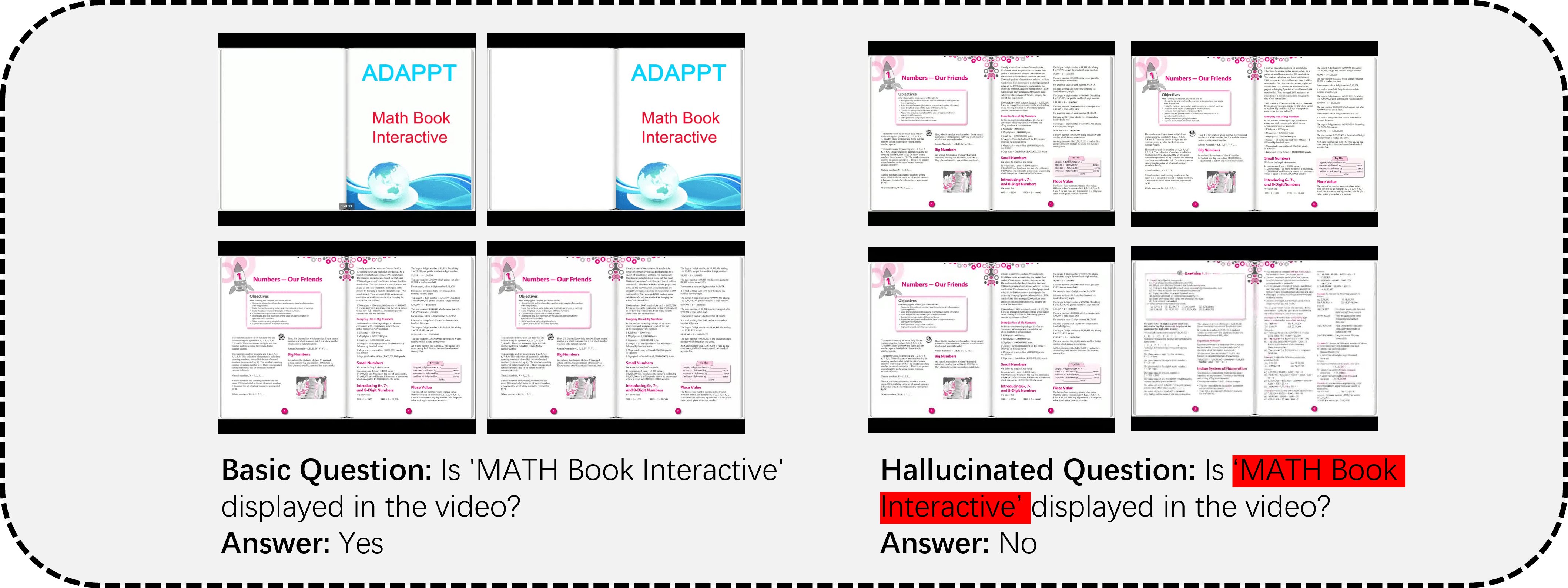}
    \caption{An example Basic-Hallucinated question pair of Semantic Detail (OCR) Hallucination}
    \label{fig:case-semantic_ocr}
\end{figure}

\begin{figure}[h]
    \centering
    \includegraphics[width=\linewidth]{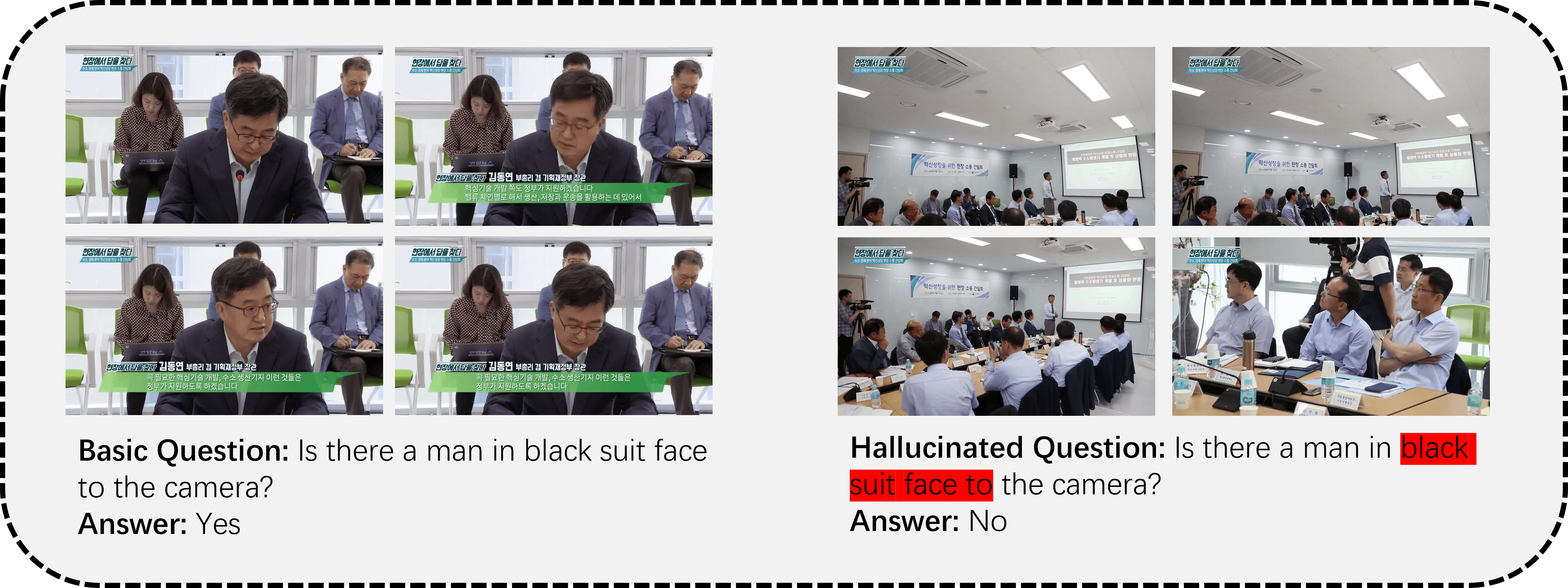}
    \caption{An example Basic-Hallucinated question pair of Semantic Detail (Camera) Hallucination}
    \label{fig:case-semantic_camera}
\end{figure}

\begin{figure}[h]
    \centering
    \includegraphics[width=\linewidth]{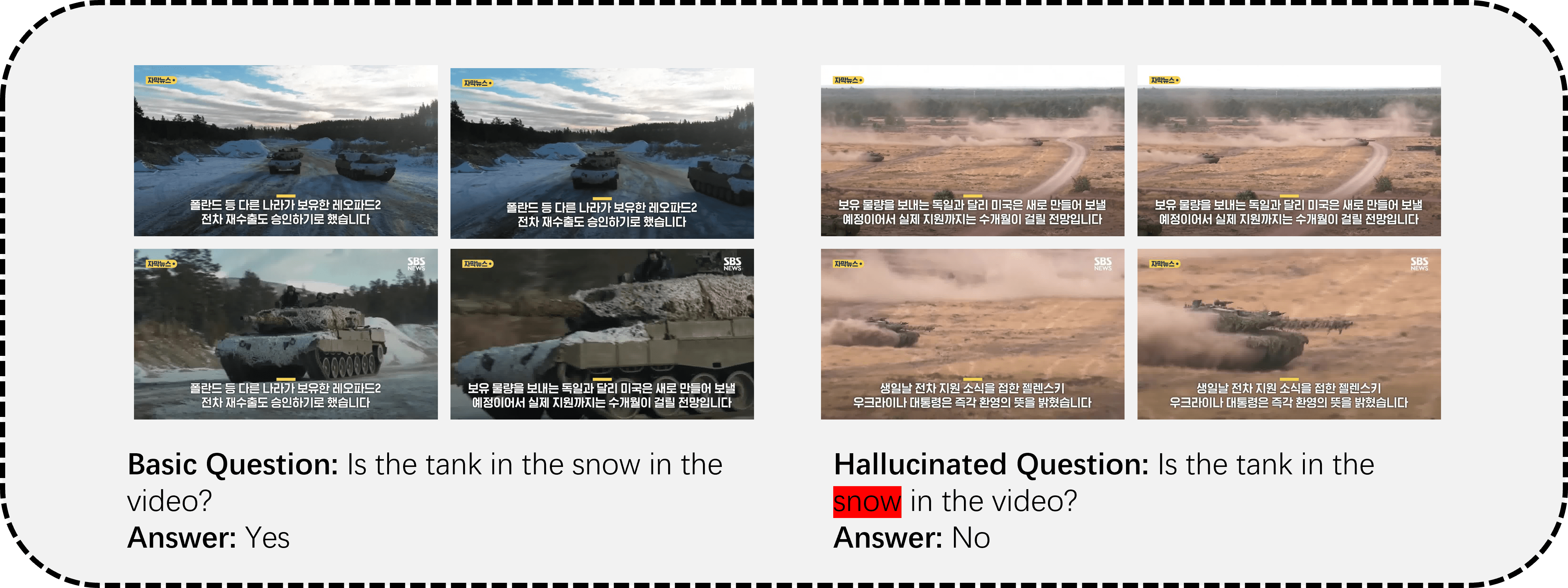}
    \caption{An example Basic-Hallucinated question pair of Semantic Detail (Scene) Hallucination}
    \label{fig:case-semantic_scene}
\end{figure}

\begin{figure}[h]
    \centering
    \includegraphics[width=\linewidth]{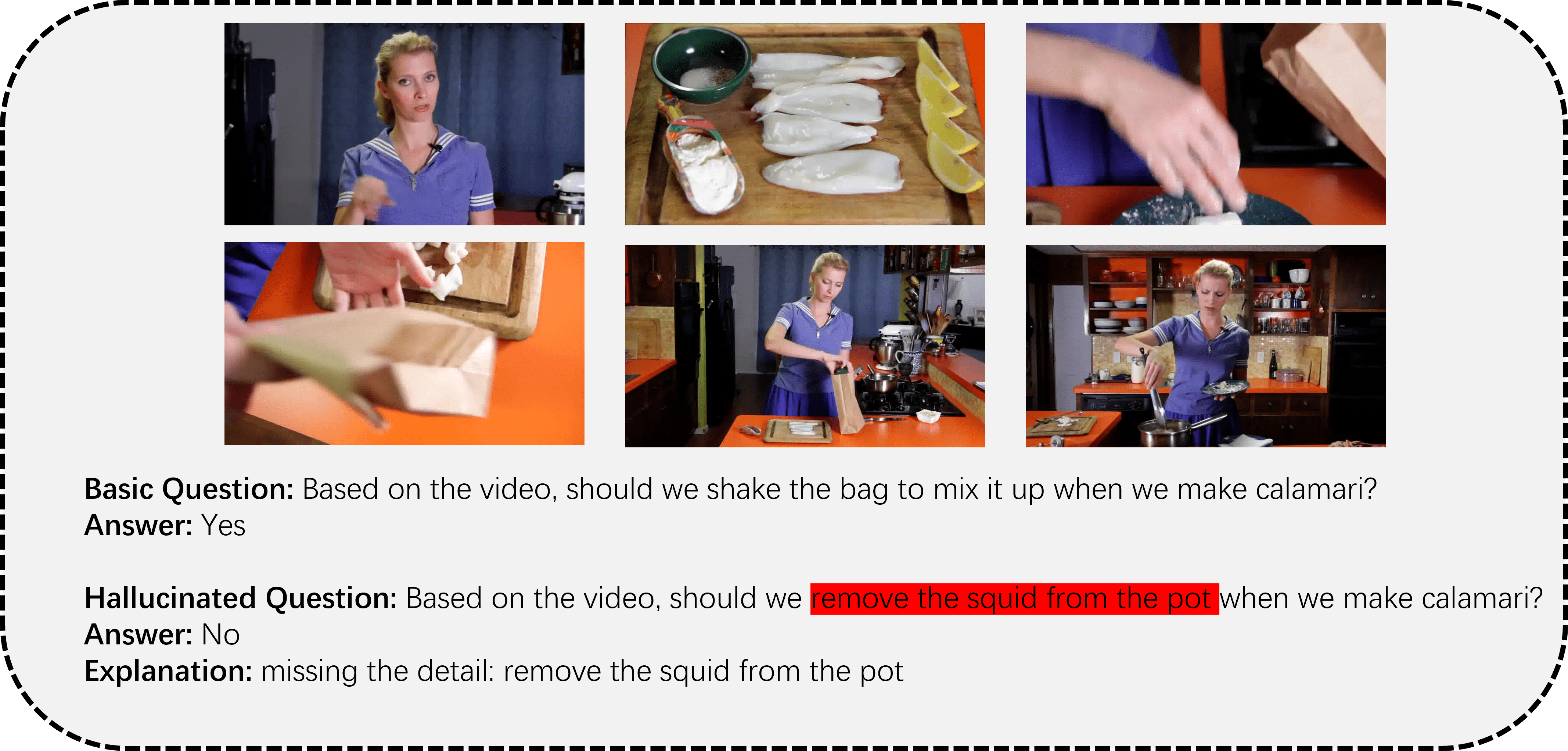}
    \caption{An example Basic-Hallucinated question pair of Extrinsic Factual (Instruction) Hallucination}
    \label{fig:case-fact_inst}
\end{figure}

\begin{figure}[h]
    \centering
    \includegraphics[width=\linewidth]{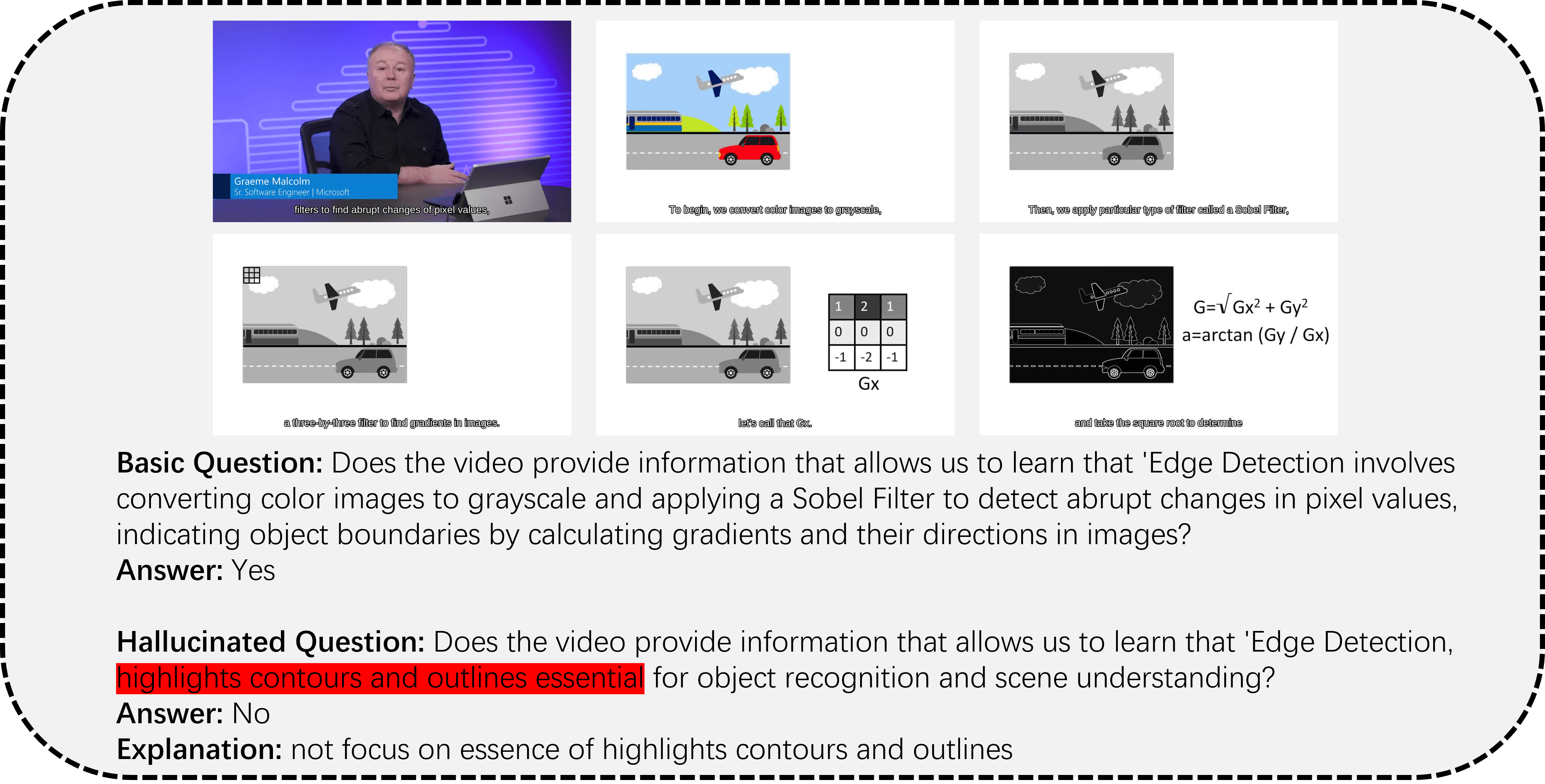}
    \caption{An example Basic-Hallucinated question pair of Extrinsic Factual (Course) Hallucination}
    \label{fig:case-fact_course}
\end{figure}

\begin{figure}[h]
    \centering
    \includegraphics[width=\linewidth]{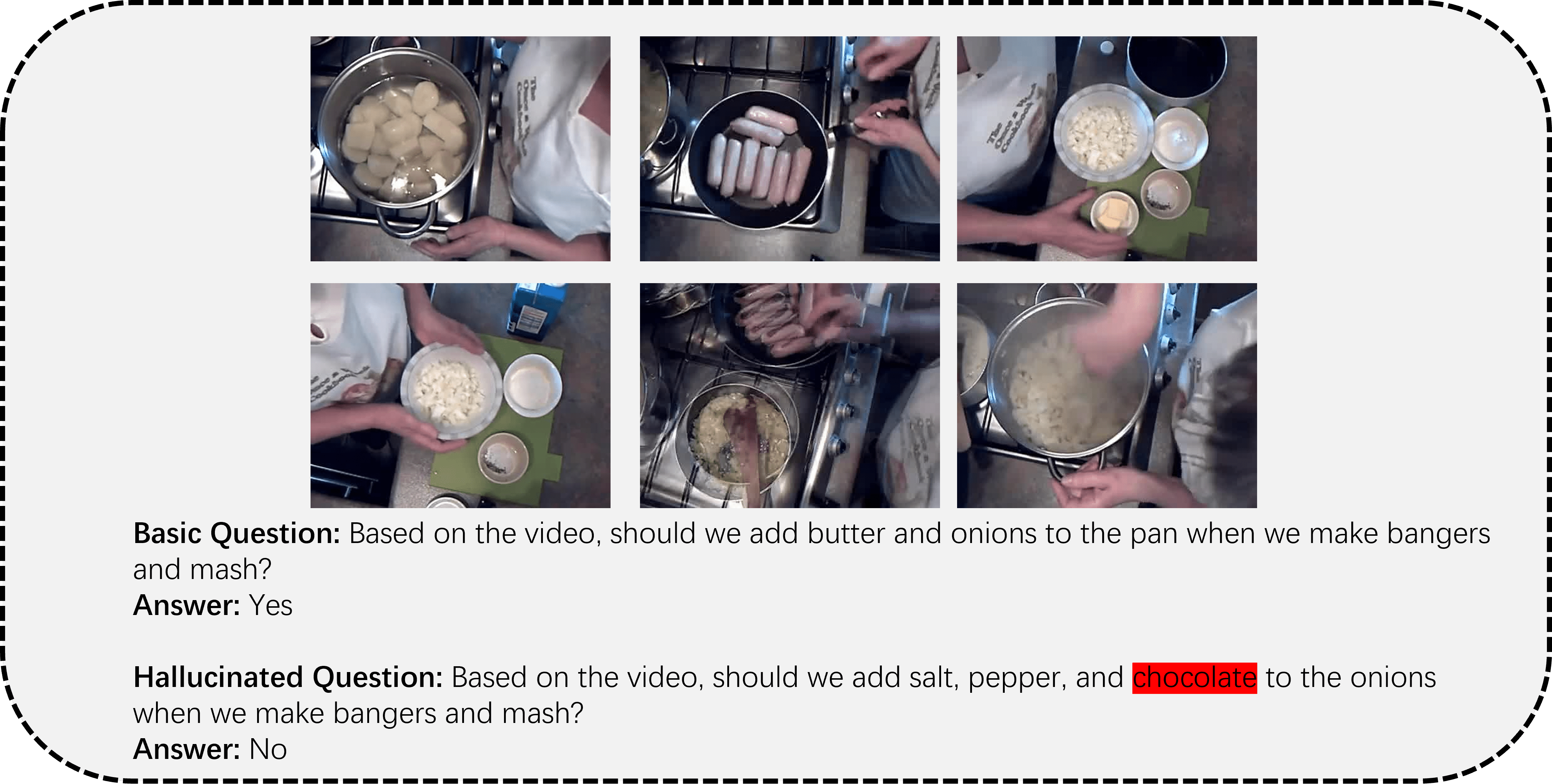}
    \caption{An example Basic-Hallucinated question pair of Extrinsic Non-factual (Instruction) Hallucination}
    \label{fig:case-nonfact_inst}
\end{figure}

\begin{figure}[h]
    \centering
    \includegraphics[width=\linewidth]{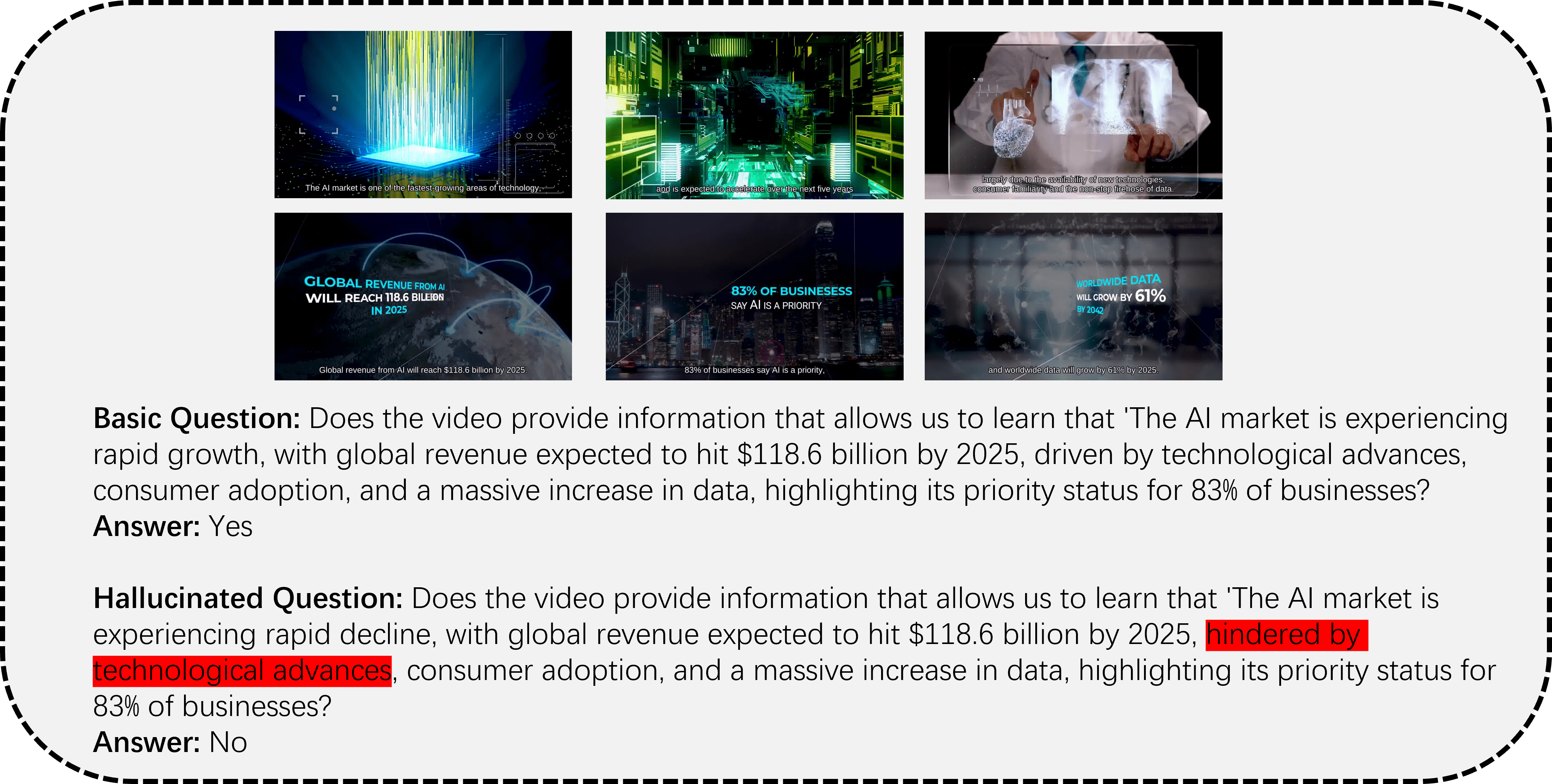}
    \caption{An example Basic-Hallucinated question pair of Extrinsic Non-factual (Course) Hallucination}
    \label{fig:case-nonfact_course}
\end{figure}

\subsection{Limitations and Ethic States}
\label{supp:limitation}

\paragraph{Limitation} Although we take a lot of strategies to make sure the quality of \bench, there is noise introduced by human annotations. In addition, currently, the scalability of this benchmark is still limited. 

\paragraph{Societal Impacts} \bench is designed to counter hallucination in LVLMs, however, the hallucinated questions in this benchmark could lead to misinterpretation in related research area. 

\paragraph{Responsibility and License}
We bear all responsibility in case of violation of rights and our dataset is under the license of CC BY-NC-SA (Attribution-NonCommercial-ShareAlike).

\end{document}